\documentclass{article} % For LaTeX2e
\usepackage{iclr2027_conference,times}

\usepackage{amsmath,amsfonts,bm}

\def\eqref#1{equation~\ref{#1}}
\def\1{\bm{1}}

\DeclareMathAlphabet{\mathsfit}{\encodingdefault}{\sfdefault}{m}{sl}
\SetMathAlphabet{\mathsfit}{bold}{\encodingdefault}{\sfdefault}{bx}{n}

\newcommand{\E}{\mathbb{E}}

\usepackage{wrapfig}
\usepackage[utf8]{inputenc} % allow utf-8 input
\usepackage[T1]{fontenc}    % use 8-bit T1 fonts
\usepackage{hyperref}       % hyperlinks
\usepackage{url}            % simple URL typesetting
\usepackage{booktabs}       % professional-quality tables
\usepackage{amsfonts}       % blackboard math symbols
\usepackage{nicefrac}       % compact symbols for 1/2, etc.
\usepackage{microtype}      % microtypography
\usepackage{xcolor}         % colors

\usepackage{amsmath,amssymb,amsfonts,mathtools,amsthm,bm}
\usepackage{booktabs,multirow,graphicx,microtype,xcolor}
\usepackage{tikz}
\usetikzlibrary{positioning,fit,calc,arrows.meta}
\usepackage{hyperref,url}
\hypersetup{hidelinks}
\usepackage{multirow}
\usepackage{algorithm}
\usepackage{algpseudocode}
\usepackage{enumitem}  % in preamble

\newtheorem{proposition}{Proposition}
\newcommand{\N}{\mathcal{N}}
\newcommand{\dd}{\mathrm{d}}
\newcommand{\FACET}{\textsc{FACET}}

\definecolor{attentionorange}{RGB}{210,95,0}

\title{FACET: \underline{F}actorized \underline{A}symmetric \underline{C}onditioning\\for \underline{E}fficient \underline{T}ransport in High-Fidelity\\Fluorescence Microscopy Synthesis}

\author{
Sazan Mahbub$^{c}$, Caleb N. Ellington$^{c,g}$, Eric P. Xing$^{c, m, g}$\\
\text{$^c$Carnegie Mellon University, $^m$Mohamed bin Zayed University of AI, $^g$GenBio AI}
}

\iclrfinalcopy % Uncomment for camera-ready version, but NOT for submission.
\begin{document}

\maketitle

\begingroup
\renewcommand\thefootnote{}
\footnotetext{Contact:
\href{mailto:smahbub@cs.cmu.edu}{smahbub@cs.cmu.edu}, \href{mailto:cellingt@cs.cmu.edu}{cellingt@cs.cmu.edu}, \href{mailto:epxing@cs.cmu.edu}{epxing@cs.cmu.edu}}
\endgroup

\begin{abstract}
Fluorescence microscopy reveals where proteins localize, but only a limited number of proteins
can be imaged in the same cell; generating these images from amino-acid sequence and the cell's
morphological context enables in silico localization of unimaged proteins. The two conditions,
however, play asymmetric roles: morphological context is spatially aligned with the target, whereas sequence is
non-spatial and must specify protein-dependent localization within it, with recurring coarse
patterns shared across proteins and finer protein-specific variation. Existing generators condition
on both jointly, without separating what each explains. We introduce FACET (\emph{Factorized Asymmetric Conditioning for Efficient Transport}), a probabilistic generative framework that encodes this
structure as an explicit inductive bias: sequence semantics are learned from what context leaves
unexplained, coarse localization regularities are shared across proteins through a semantic
memory, and protein-specific variation is a bounded residual around them. A variance-preserving
state projection further lets FACET perform continuous stochastic transport through a pretrained
diffusion predictor with minimal parameter overhead. On held-out proteins, FACET improves spatial
overlap by 34.3\% on the Human Protein Atlas and 14.0\% on OpenCell over a backbone-matched
baseline, and reduces FID by 27.2\% and 46.5\%, respectively, with 75\% fewer network evaluations.
It also substantially improves protein-association structure recovery and yields better-calibrated predictions,
while detailed ablations show complementary contributions from its  design choices.
These results identify factorized asymmetric conditioning, rather than generator capacity alone,
as a key lever for high-fidelity, efficient, and biologically meaningful cellular image synthesis.
\end{abstract}

\vspace{-1em}
\section{Introduction}
\label{sec:intro}
\vspace{-1em}

Protein subcellular localization is a spatial phenotype {that links molecular
identity to cellular function and is directly observed} through fluorescence
microscopy~\citep{thul2017hpa}. Recent sequence-to-image models generate these phenotypes from
protein sequence and cellular context~\citep{khwaja2023celle2,zheng2025celldiff,zhang2025prediction}. Yet these
conditions play fundamentally different roles: context is spatially aligned with the target,
whereas sequence is non-spatial and must specify protein-dependent localization semantics within
that context. The conditions are therefore asymmetric in representation and explanatory role,
rather than in biological importance.

\textbf{Our key insight} is that protein localization is inherently
context-relative and hierarchically organized, spanning from recurring regularities shared across
proteins to finer protein-specific variation. This motivates three coupled
modeling requirements. First, sequence representations should capture localization information beyond what context
already explains: even a linear map from context features explains over a third of held-out
target-feature variance (Appendix~\ref{app:prototypes}), a lower bound on the redundancy that
sequence semantics need not re-encode. Second, recurring coarse localization regularities
should be shared across proteins rather than relearned independently. Third, finer
protein-specific variation should remain distinct from both the shared regularities and image
stochasticity.

We introduce \FACET{}, a
probabilistic generative framework that translates this hierarchy into asymmetric conditioning:
sequence semantics are factorized into coarse semantics, retrieved from a memory built on
context-residualized features, and finer protein-specific variation, which jointly condition
stochastic transport through a pretrained diffusion model. This formulation encodes an \textit{explicit
inductive bias}: sequence semantics are learned from what context leaves unexplained, coarse localization
regularities are shared across proteins, and protein-specific variation is bounded relative to
them, a structure that favors generalization to unseen proteins.

% We introduce \FACET{} (\emph{Factorized Asymmetric Conditioning for Efficient Transport}), a
% probabilistic generative framework that translates this hierarchy into asymmetric conditioning.
% To avoid redundantly encoding spatial information already explained by context, \FACET{} builds a semantic memory from context-residualized target features. To exploit shared
% localization regularities without sacrificing protein specificity, sequence semantics are
% factorized into coarse semantics captured by this memory and finer protein-specific variation.
% This factorization conditions a stochastic transport process that efficiently
% reuses a pretrained diffusion prior. This formulation encodes an explicit inductive bias: sequence explains only what context leaves
% unexplained, enables re-usage of shared coarse semantics of localization regularities, and protein-specific variation is bounded relative to these shared semantics, a
% structure well suited to generalizing to unseen proteins.

\FACET{} improves spatial overlap by $34.3\%$ on the Human Protein Atlas (HPA;~\citealp{thul2017hpa}) and
$14.0\%$ on OpenCell~\citep{cho2022opencell} benchmarks over a backbone-matched baseline, and reduces FID by $27.2\%$ and $46.5\%$,
respectively, with $75\%$ fewer network evaluations. It also strengthens protein-association separation by $39$--$87\%$ across association-confidence
thresholds and reduces calibration error by $74.7\%$.

Our major contributions are:
% \begin{itemize}
% \begin{itemize}[leftmargin=*, itemsep=0.2ex, topsep=-1pt]
\begin{itemize}[leftmargin=*, itemsep=0.2pt, topsep=-1pt]

\item \emph{Factorized asymmetric conditioning.}
We introduce \FACET{}, a generative framework that assigns cellular context, shared localization
semantics, protein-specific variation, and image stochasticity to separate conditioning roles,
using a context-residualized semantic memory while preserving their hierarchical dependencies.

\item \textit{Principled latent-variable formulation with efficient transport.} We derive a conditional latent-variable model whose evidence lower bound decomposes into
coarse-semantic, protein-specific, and transport terms, each optimized blockwise by our staged
objective, and establish an exact variance-preserving state projection that enables continuous
stochastic transport through a pretrained finite-schedule latent diffusion predictor with under
0.6\% parameter overhead.

% \item \attention{\emph{Theory-grounded framework for efficient stochastic transport.}
% We derive a conditional latent-variable formulation and ELBO for the semantic
% factorization, and establish variance-preserving state compatibility that enables
% continuous stochastic transport to leverage a pretrained finite-schedule
% diffusion prior with less than \(0.6\%\) parameter overhead.}

\item \emph{Broad empirical gains and controlled analyses.}
On HPA and OpenCell, \FACET{} substantially improves spatial agreement and distributional fidelity with
$75\%$ fewer network evaluations. Controlled ablations
isolate the contributions of its semantic, hierarchical, and transport components, while further
analyses show cross-dataset transfer, consistent gains across multi-channel
context compositions, stronger recovery of protein-association patterns, and improved predictive
calibration.

% \item \emph{Broad empirical gains and controlled analyses.}
% Across HPA and OpenCell, \FACET{} improves spatial agreement and distributional
% fidelity with $75\%$ fewer network evaluations. Controlled ablations provide
% direct evidence for the contributions of the proposed design choices, while
% further analyses demonstrate cross-dataset transfer, robustness across context
% compositions, stronger recovery of protein-association patterns, and improved predictive
% calibration.

\end{itemize}

\paragraph{Related work.} Sequence-conditioned localization imaging has progressed from
bidirectional sequence-image transformers~\citep{khwaja2023celle2} and sequence- and landmark-conditioned prediction~\citep{zhang2025prediction} to diffusion conditioned on sequence and
morphology~\citep{zheng2025celldiff}; concurrent preprints scale identity-conditioned generation to 12,800 human proteins~\citep{proticelli} or adopt autoregressive
generation~\citep{kambhampati2026vermeer}. These works mainly advance the image model;
FACET instead factorizes sequence semantics and cellular context within a conditional
generator, complementing any pretrained variance-preserving backbone, so we compare in a
backbone-matched setting. Extended discussion on related work is provided in Appendix~\ref{app:related}.

\section{FACET: Factorized Asymmetric Conditioning for Efficient Transport}
\label{sec:method}

\subsection{Factorized probabilistic model}
\label{sec:probabilistic-model}

\paragraph{Modeling objective.}
% We model the target image distribution \(p_\Theta(Y\mid S,C)\), conditioned on protein sequence \(S\) and cellular morphology \(C\). 
% Let $S$, $C$, and $Y$ denote the protein sequence, cellular morphology context,
% and target image, respectively. Our objective is to model the conditional target
% distribution $p_\Theta(Y\mid S,C)$.
We study conditional protein spatial-phenotype generation, where a target image $Y$ is generated from protein sequence $S$ and cellular morphology context $C$.
While \(C\) specifies the
spatial environment, sequence-dependent localization spans coarse regularities
shared across proteins and finer protein-specific variation. We represent this hierarchy with a reusable semantic memory $\mathcal M=\{V_r\}_{r=1}^{K}$ of
coarse localization prototypes, a prototype-valued latent $R$ for
sequence-dependent prototype selection, and a continuous residual $\delta z_S$
for variation beyond the prototypes. This
factorization organizes coarse and fine sequence variation while retaining direct
conditioning on both \(S\) and \(C\) in the image model.

% \paragraph{Latent variables and fixed semantic memory.}
With target image encoding $z_1=E_Y(Y)$, we model
\begin{equation}
p_\Theta(Y\mid S,C,\mathcal M)
= \sum_R \int p(Y\mid z_1)\,
p_\Theta(z_1,z_0,R,\delta z_S\mid S,C,\mathcal M)\,
dz_1\,dz_0\,d\delta z_S,
\label{eq:conditional-object}
\end{equation}
% where $z_1=E(Y)$ encodes $Y$ and $\mathcal M=\{V_r\}_{r=1}^{K}$ is the reusable prototype memory.

\paragraph{Latent-variable factorization.}
With a fixed, deterministic decoder $D$, we take
$p(Y\mid z_1)=\delta\!\left(Y-D(z_1)\right)$,
so learning the conditional image distribution reduces to modeling
the latent distribution. We factorize the latter as
% \begin{equation}
% % \begin{aligned}
%  p_\Theta(z_1,z_0,R,\delta z_S\mid S,C,\mathcal M)
%  =p_0(z_0)
%  p_\eta(R\mid S,\mathcal M)
%  p_\psi(\delta z_S\mid S,R)
%  p_\theta(z_1\mid z_0,S,C,R,\delta z_S).
% % \end{aligned}
% \label{eq:factor_full}
% \end{equation}
\begin{equation}
\begin{aligned}
p_\Theta(z_1,z_0,R,\delta z_S\mid S,C,\mathcal M)\hspace{0pt}=\hspace{-8pt}
\underbrace{p_0(z_0)}_{\text{base distribution}}\hspace{0pt}
\underbrace{p_\eta(R\mid S,\mathcal M)}_{\text{prototype prior}}
\,\underbrace{p_\psi(\delta z_S\mid S,R)}_{\text{residual prior}}
\,\underbrace{p_\theta(z_1\mid z_0,S,C,R,\delta z_S)}
_{\text{conditional transport}}.
\end{aligned}
\label{eq:factorization}
\end{equation}

% \begin{equation}
% \begin{aligned}
%  p_\Theta(z_1,z_0,R,\delta z_S\mid S,C,\mathcal M)
%  ={}&p_0(z_0)
%  p_\eta(R\mid S,\mathcal M)
%  p_\psi(\delta z_S\mid S,R)\\
%  &\qquad\qquad\times
%  p_\theta(z_1\mid z_0,S,C,R,\delta z_S).
% \end{aligned}
% \label{eq:factor_full}
% \end{equation}
Here, $\eta$ and $\psi$ parameterize the prototype and residual priors,
respectively. After integrating out the initial latent $z_0$, this becomes
\begin{equation}
 p_\Theta(z_1,R,\delta z_S\mid S,C,\mathcal M)
 =p_\eta(R\mid S,\mathcal M)
 \,p_\psi(\delta z_S\mid S,R)
 \,p_\theta(z_1\mid S,C,R,\delta z_S).
\label{eq:factor}
\end{equation}
\newcommand{\pgmscale}{0.98}

\begin{wrapfigure}{r}{0.45\textwidth}
\vspace{-15pt}
\centering

\resizebox{\pgmscale\linewidth}{!}{%
\begin{tikzpicture}[
yscale=0.60,
xscale=0.55,
font=\small,
>=Latex,
obs/.style={
    circle,
    draw,
    line width=0.7pt,
    fill=gray!18,
    minimum size=9mm,
    inner sep=1pt
},
lat/.style={
    circle,
    draw,
    line width=0.7pt,
    minimum size=9mm,
    inner sep=1pt
},
fix/.style={
    draw,
    line width=0.7pt,
    rounded corners=3pt,
    fill=gray!10,
    minimum height=7.5mm,
    minimum width=9mm,
    inner sep=3pt
},
arr/.style={
    ->,
    line width=0.85pt,
    shorten >=2pt,
    shorten <=2pt
}
]

\node[fix] (M)  at (0,3.9)     {$\mathcal M$};
\node[obs] (S)  at (0,1.6)     {$S$};
\node[obs] (C)  at (0,-0.1)    {$C$};
\node[lat] (R)  at (3.0,3.9)   {$R$};
\node[lat,minimum size=11mm] (D) at (5.6,1.7) {$\delta z_S$};
\node[lat] (Z0) at (8.4,3.9)   {$z_0$};
\node[lat] (Z1) at (8.4,-0.1)  {$z_1$};
\node[obs] (Y)  at (11.4,-0.1) {$Y$};

\draw[arr] (M)  -- (R);
\draw[arr] (S)  -- (R);
\draw[arr] (S)  -- (D);
\draw[arr] (R)  -- (D);
\draw[arr] (D)  -- (Z1);
\draw[arr] (R)  to[bend left=40] (Z1);
\draw[arr] (S)  -- (Z1);
\draw[arr] (C)  -- (Z1);
\draw[arr] (Z0) -- (Z1);
\draw[arr] (Z1) -- (Y);

\node[
    anchor=west,
    align=left,
    font=\scriptsize
] at (9.1,2.0) {
    observed: $S,C,Y$\\
    latent: $R,\delta z_S,z_0,z_1$\\
    fixed memory: $\mathcal M$\\
    training-time inference\\
    paths omitted
};

\end{tikzpicture}%
}
\vspace{-1em}
\caption{Probabilistic graphical model for \FACET{}.}
\label{fig:pgm}
\vspace{-1em}
\end{wrapfigure}
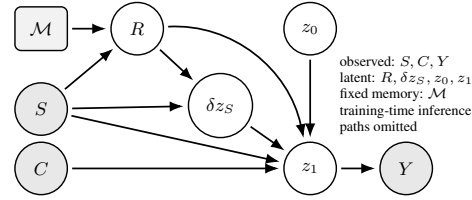

%
% Marginalizing $R$ and $\delta z_S$ further yields $p_\Theta(z_1\mid S,C,\mathcal M)$ in Eq.~\ref{eq:conditional-object}. 
Since $R$ is prototype-valued, its selected memory state directly conditions
the residual and transport factors.

These roles are deliberately asymmetric: $C$ explains cell-specific spatial
organization, $R$ supplies coarse sequence-dependent semantics from $\mathcal M$,
$\delta z_S$ captures fine-grained variations beyond the prototypes, and
$p_\theta$ models realization-level image variation. 
% This assigns context and
% sequence semantics distinct roles, while context-residualized memory reduces
% representational redundancy between them.
% This decomposition reduces redundancy between the strong spatial
% context and the sequence-dependent semantic state while retaining both $S$ and
% $C$ as direct conditions of the target latent.

% \paragraph{Evidence lower bound.} The factorization induces the following conditional evidence lower bound. 
\vspace{.5em}
\begin{proposition}[Conditional evidence lower bound]\label{prop:elbo}
For any variational posterior
$
q(R,\delta z_S\mid z_1,S,C,\mathcal M)
=q_\omega(R\mid z_1,S,C,\mathcal M)
 \,q_\phi(\delta z_S\mid z_1,S,C,R,\mathcal M),
$
we have
\begin{equation}
\begin{aligned}
\log p_\Theta(z_1\mid S,C,\mathcal M)
\hspace{-2pt}\geq{}\hspace{-1pt}&
\mathbb E_q\left[\log p_\theta(z_1\mid S,C,R,\delta z_S)\right]
\hspace{-2pt}-\hspace{-2pt}D_{\mathrm{KL}}\!\left(
q_\omega(R\mid z_1,S,C,\mathcal M)
\,\middle\|\,
p_\eta(R\mid S,\mathcal M)
\right)\\
&-\mathbb E_{q_\omega}D_{\mathrm{KL}}\!\left(
q_\phi(\delta z_S\mid z_1,S,C,R,\mathcal M)
\,\middle\|\,
p_\psi(\delta z_S\mid S,R)
\right).
\end{aligned}
\label{eq:elbo}
\end{equation}
\end{proposition}

\vspace{-8pt}
Here $D_{\mathrm{KL}}$ is the Kullback-Leibler divergence, and $\omega,\phi$
parameterize the amortized posteriors. 
% \revnew{The bound probabilistically formalizes the factorized model}; 
See Appendix~\ref{app:proof-elbo} for proof.

\subsection{Context-residualized semantic memory}
\label{sec:semantic-memory}

\paragraph{Context residualization.}
To avoid relearning spatial information already available from $C$, we construct memory from
context-residualized target features. With fixed features $\Phi_Y(Y_i)$ and
$\Phi_C(C_i)$, an affine ridge predictor $b$ fit on training data defines
$U_i=\Phi_Y(Y_i)-b(\Phi_C(C_i))$, retaining target features unexplained by
context. A fitted $T_C$ compresses the context prediction as
$z_C=f_C(C)=T_C(b(\Phi_C(C)))$. Further details and held-out explanatory analysis are
in Appendix~\ref{app:prototypes}.
% To avoid relearning spatial information already available from $C$, we construct the memory in a context-residualized target-feature space. For fixed target and context features $\Phi_Y(Y_i)$ and $\Phi_C(C_i)$, we fit an affine ridge predictor $b$ on training data and define $ U_i=\Phi_Y(Y_i)-b\!\left(\Phi_C(C_i)\right),$ retaining target features unexplained by context. We further compress the context prediction with a fitted transformation $T_C$: $z_C=f_C(C)=T_C(b(\Phi_C(C)))$. Additional details and held-out context explanatory analysis are reported in Appendix~\ref{app:prototypes}.

% \paragraph{Context residualization.}
% A sequence-dependent representation need not relearn spatial structure that is
% already available from $C$. We therefore construct the reusable memory in a
% target-feature residual space. Let $\Phi_Y(Y_i)$ and $\Phi_C(C_i)$ denote fixed
% target and context features for training image $i$. Using training data only,
% we fit an affine ridge predictor $b$ from context features to target features
% and form
% \begin{equation}
% U_i=\Phi_Y(Y_i)-b\!\left(\Phi_C(C_i)\right).
% \label{eq:resid_per_sample}
% \end{equation}
% The residual retains target features not predicted
% by the fitted context map. The same ridge prediction $b(\Phi_C(C))$ is also
% compressed by a fitted transformation $T_C$ to give
% $z_C=f_C(C)=T_C(b(\Phi_C(C)))$. The corresponding held-out context
% explanatory analysis is reported in Appendix~\ref{app:prototypes}.

\paragraph{Protein-level prototype construction.}
For each training protein $p$, we average residuals over its images $\mathcal I_p$
and apply a fixed standardized PCA projection $T$ as
$%\begin{equation}
e_p=T\!\left(
\frac{1}{|\mathcal I_p|}\sum_{i\in\mathcal I_p}U_i
\right),
% \label{eq:protein-prototype-representation}
$ %\end{equation}
giving proteins equal weight and a compact shared representation. A fixed
annotation-derived rule assigns each protein to one of $K$ coarse semantic
states; for proteins $\mathcal P_k$ assigned to state $k$, we define the centroid
$%\begin{equation}
V_k=\frac{1}{|\mathcal P_k|}
\sum_{p\in\mathcal P_k}e_p \text{ and memory }
\mathcal M=\{V_k\}_{k=1}^{K}.
% \label{eq:prototype-centroids}
$ %\end{equation}
 The projection, assignments, and centroids use training proteins only and are
then frozen; construction details and dimensions are in
Appendix~\ref{app:prototypes}.

% \paragraph{Protein-level prototype construction.}
% Because the memory represents proteins rather than individual images, we first
% average residual features across the images $\mathcal I_p$ of each training
% protein $p$ and then apply a fixed standardized principal-component projection
% $T$:
% \begin{equation}
% e_p=T\!\left(
% \frac{1}{|\mathcal I_p|}\sum_{i\in\mathcal I_p}U_i
% \right).
% \label{eq:protein-prototype-representation}
% \end{equation}
% This gives every protein equal weight regardless of its number of images and
% places the memory elements in a compact shared coordinate system. A fixed
% annotation-derived rule assigns each memory-construction protein to one of $K$
% coarse semantic states. If $\mathcal P_k$ denotes the proteins assigned to
% state $k$, its prototype is the centroid
% \begin{equation}
% V_k=\frac{1}{|\mathcal P_k|}
% \sum_{p\in\mathcal P_k}e_p,
% \qquad
% \mathcal M=\{V_k\}_{k=1}^{K}.
% \label{eq:prototype-centroids}
% \end{equation}
% The transformations, assignments, and centroids are estimated using training
% proteins and then frozen. Additional construction details and dimensions are
% provided in Appendix~\ref{app:prototypes}.

\paragraph{Sequence-conditioned soft memory readout.}
At inference, target images and annotation-derived states are unavailable; the
sequence prior instead yields the soft
readout
$
m_S=\E[R\mid S,\mathcal M]
=\sum_{r=1}^{K}p_\eta(R=V_r\mid S,\mathcal M)V_r.
$
This inference-time approximation preserves graded semantics without discrete sampling
noise. 

\vspace{-.2em}
\subsection{Bounded variational sequence residual}
\label{sec:variational-residual}
\vspace{-.2em}
% Context residualization shapes the coarse memory $\mathcal M$, while 
The residual $\delta z_S$ captures fine-grained protein-specific
variation beyond the coarse readout $m_S$.

\vspace{-.4em}
\paragraph{Target-informed posterior and sequence-only prior.}
We define the refined semantic state as
$
z_S=m_S+\delta z_S
% \label{eq:semantic-state}
$
and protein language model encoding as $h_S=E_S(S)$. 
For each training image, the posterior input is
the projected residual $U_Y=T(\Phi_Y(Y)-b(\Phi_C(C)))$.
The target-informed amortized posterior and sequence-only prior are
$
q_\phi(\delta z_S\mid U_Y,m_S)
=\N\!\left(\mu_q(U_Y,m_S),\sigma_z^2I\right)$ 
and 
$
p_\psi(\delta z_S\mid h_S,m_S)
=\N\!\left(\mu_p(h_S,m_S),\sigma_z^2I\right),
$
respectively.

% \begin{align}
% q_\phi(\delta z_S\mid U_Y,m_S)
% &=\N\!\left(\mu_q(U_Y,m_S),\sigma_z^2I\right),\\
% p_\psi(\delta z_S\mid h_S,m_S)
% &=\N\!\left(\mu_p(h_S,m_S),\sigma_z^2I\right).
% \label{eq:vi}
% \end{align}
% The posterior can use target information during training, whereas only the sequence-conditioned prior is available during generation.

\paragraph{Fixed-scale divergence and bounded parameterization.}
Both distributions use the same fixed isotropic covariance. Consequently,
\vspace{-.2em}
\begin{equation}
\vspace{-.2em}
D_{\mathrm{KL}}(q_\phi\|p_\psi)
=\frac{\|\mu_q-\mu_p\|_2^2}{2\sigma_z^2}.
\label{eq:fixedkl}
\vspace{-.2em}
\end{equation}
The fixed scale prevents the optimizer from reducing the divergence by inflating
a learned variance (Appendix~\ref{app:proof-fixed-kl}). We parameterize the residual scale relative to the empirical
coarse-memory magnitude $s_m$ as $\sigma_z=c_\sigma s_m$, and bound the
posterior and prior means using
$
\mu_a=c_\mu s_m\tanh f_a(\cdot),
~ c_\sigma,c_\mu>0, a\in\{q,p\}.
% \label{eq:bounded-mean}
$
Thus, the fine residual has an explicit scale relative to the coarse semantic
state. The implementation details are reported in Appendix~\ref{app:setup}.

% \paragraph{Inference-time semantic state.}
% For the default point-estimate experiments, we use the prior mean
% $
% z_S=m_S+\mu_p(h_S,m_S).
% % \label{eq:prior-mean-state}
% $
% Predictive-sampling experiments instead draw the protein-specific residual from its prior using the reparameterization
% $
% z_S=m_S+\mu_p(h_S,m_S)+\sigma_z\,\varepsilon, \text{ with } \varepsilon\sim\N(0,I).
% % \label{eq:prior-reparameterized-sampling-state}
% $
% % $$\delta z_S\gets\mu_p(h_S,m_S)+\sigma_z,\varepsilon,\qquad \varepsilon\sim\N(0,I)$$
% % $$z_S\gets m_S+\delta z_S.$$
% %
% % $$z_S=m_S+\delta z_S,\qquad \delta z_S\sim p_\psi(\delta z_S\mid h_S,m_S)=\N!\left(\mu_p(h_S,m_S),\sigma_z^2I\right),$$
% % i.e. reparameterized, with $\sigma_z=c_\sigma s_m$:
% % $$z_S=m_S+\mu_p(h_S,m_S)+\sigma_z,\varepsilon,\qquad \varepsilon\sim\N(0,I).$$

\vspace{-.2em}
\subsection{Asymmetric stochastic transport}
\label{sec:transport}
\vspace{-.2em}
\subsubsection{Variance-preserving stochastic interpolant}
\label{sec:interpolant}
\vspace{-.2em}
\paragraph{Stochastic interpolant.}
We learn continuous stochastic transport from $z_0\sim\N(0,I)$ to the target
latent $z_1$. {For $\tau\in[0,1]$, define the
stochastic interpolant~\citep{albergo2023stochastic}}
\begin{equation}
x_\tau=A_\tau z_1+B_\tau z_0+G_\tau\xi,
\qquad
A_\tau=\tau,\quad B_\tau=1-\tau,\quad
G_\tau=\kappa\tau(1-\tau),
\label{eq:si}
\end{equation}
where $\xi\sim\N(0,I)$ and $\kappa>0$. Defining
$Q_\tau=\sqrt{B_\tau^2+G_\tau^2}$ gives
\begin{equation*}
x_\tau=A_\tau z_1+Q_\tau\epsilon_\tau,
\qquad \epsilon_\tau\sim\N(0,I).
\end{equation*}

\paragraph{Variance-preserving normalization.}
To leverage the learned image diffusion prior of a pretrained $\varepsilon$-predictor trained on a finite variance-preserving schedule,
we normalize each continuous interpolant state
$x_\tau$ onto the same variance-preserving family.

\begin{proposition}[Variance-preserving state compatibility]
\label{prop:vp}
Let $N_\tau=\sqrt{A_\tau^2+Q_\tau^2}$ and $y_\tau=x_\tau/N_\tau$. Then
\begin{equation}
y_\tau=a_\tau z_1+\sigma_\tau\epsilon_\tau,
\qquad
a_\tau=\frac{A_\tau}{N_\tau},
\quad
\sigma_\tau=\frac{Q_\tau}{N_\tau},
\label{eq:vp-state}
\end{equation}
with $a_\tau^2+\sigma_\tau^2=1$ exactly.
\end{proposition}

% The state projection is exact. Compatibility with the finite pretrained schedule
% is approximated through nearest-index lookup:
Proposition~\ref{prop:vp} folds both noise sources into a single effective noise, so a frozen
pretrained $\varepsilon$-predictor can drive continuous stochastic transport with minimal
parameter overhead. Unlike Diff2Flow \citep{schusterbauer2025diff2flow}, which targets
deterministic flow matching, it also covers the latent noise $G_\tau\xi$. The projection is
exact; only the finite-schedule lookup is approximate:
% The state projection is exact. The remaining approximation is nearest-index
% lookup on the finite native schedule, with
\begin{equation}
k(\tau)=\arg\min_k|\tau_k-\tau|.
\label{eq:schedule-lookup}
\end{equation}
The schedule-specific representability details are provided in
Appendix~\ref{app:flow}.

\paragraph{Base velocity.}
% Given the adapted $\varepsilon$-predictor output $\hat\epsilon_L$, {which incorporates the parameter-efficient updates and $\varepsilon$-adapter,} we form a base clean estimate and velocity:
% The $\varepsilon$-adapter is a lightweight module that adds an $m_S$-conditioned residual
% correction to the pretrained noise prediction; with LoRA, this yields the adapted noise
% estimate $\hat{\varepsilon}_L$, from which we form a base clean estimate and velocity:
Given the adapted noise estimate $\hat{\epsilon}_L$, i.e., the LoRA-adapted pretrained
prediction followed by an $\varepsilon$-adapter, a lightweight module that adds an $m_S$-conditioned
residual correction, we form a base clean estimate and velocity:
\begin{equation}
\hat z_1^L
=\frac{x_\tau-Q_\tau\hat\epsilon_L}
{\max(A_\tau,\epsilon_{\rm num})},
\qquad
v_L=\hat z_1^L+\dot Q_\tau\hat\epsilon_L.
\label{eq:base-velocity}
\end{equation}
See Appendix~\ref{app:base-velocity-derivation} for the full derivation and further details.

\subsubsection{Context-first hierarchical correction}
\label{sec:hierarchical-correction}

\paragraph{Ordered modulation.}
Let $H_k$ be an intermediate image feature at native noise index $k$ and
$e_\tau$ a continuous-time embedding. FACET applies feature-wise linear
modulation (FiLM)~\citep{perez2018film} in an ordered correction:
\begin{equation}
H_C=\mathrm{FiLM}_C(H_k,[z_C;e_\tau]),
\qquad
H_{CS}=\mathrm{FiLM}_S(H_C,[z_S;e_\tau]).
\label{eq:film}
\end{equation}
The sequence representation therefore refines a feature representation that has
already absorbed cellular context rather than competing with context at the same
injection point.

\paragraph{Gated sequence short-cut.}
A sequence short-cut and a lightweight residual correction head produce the
velocity correction $\Delta v$:
\begin{equation}
H_{CS}\leftarrow H_{CS}+g(\tau)P_z(z_S),
\qquad
g(\tau)=4\tau(1-\tau).
\label{eq:gating}
\end{equation}
Here $P_z$ projects the protein state to a channel-wise bias broadcast over the
feature map. The correction head is zero-initialized so that the added path is
inactive at initialization. The gate vanishes at both transport endpoints and
has unit magnitude at the midpoint.

\vspace{-.5em}
\paragraph{Role-specific conditioning paths.}
The resulting computation preserves the pretrained predictor's direct sequence
input, uses $m_S$ in the $\varepsilon$-adapter, {uses $E_C(C)$ in the backbone and the compact context state
$z_C=f_C(C)$ in the first FiLM block,} and uses the refined state $z_S$
only in the hierarchical sequence correction. This allocation allows the pretrained predictor to retain
its original conditional pathway while the learned modules provide structured
sequence-dependent refinement.

\vspace{-.5em}
\paragraph{Velocity composition and parameter-efficient adaptation.}
The final velocity field is
\begin{equation}
v_\Theta(x_\tau,\tau,S,C)
=v_L(x_\tau,\tau,S,C)
+g(\tau)\Delta v(H_k,z_C,z_S,\tau).
\label{eq:velocity}
\end{equation}
The original image-predictor weights, VAE, and protein encoder remain frozen.
Low-Rank Adaptation (LoRA)~\citep{hu2022lora} supplies parameter-efficient
updates to the $\varepsilon$-predictor, while the $\varepsilon$-adapter,
semantic prior, continuous prior, and correction head are trained. The exact
module configuration and parameter counts are provided in Appendix~\ref{app:setup}.

\vspace{-.5em}
\subsection{Training and inference}
\label{sec:training-inference}
\vspace{-.5em}

\paragraph{Modular training.}
% %
% \input{figures/qualitative_main_text} 
% %
Rather than optimizing the ELBO jointly, \FACET{} follows a staged
procedure aligned with its factorization: we establish the context-residualized
memory and context map, pretrain and freeze the prototype prior and
memory-conditioned $\varepsilon$-adapter, and then optimize the continuous
residual and conditional transport components. The transport-stage objective
combines conditional flow matching~\citep{lipman2023flow} with reconstruction,
sequence-ranking, variational, and residual regularization:
\begin{align}
\vspace{-.5em}
\mathcal L={}&\mathcal L_{\rm FM}
+\lambda_{\rm rec}\mathcal L_{\rm rec}
+\lambda_{\rm pix}\mathcal L_{\rm rec,pix}
+\lambda_{\rm rank}\mathcal L_{\rm rank}
+\lambda_{\rm VI}D_{\rm KL}(q_\phi\|p_\psi)\nonumber\\
&+\lambda_{\rm shrink}\mathcal L_{\rm shrink}
+\lambda_{\rm same}\mathcal L_{\rm same}.
\label{eq:transport-objective}
\vspace{-.7em}
\end{align}
% %
% \input{figures/qualitative_main_text} 
% %
%%%%
This staged procedure optimizes the terms of Eq.~\ref{eq:elbo} blockwise: $\mathcal{L}_{\mathrm{FM}}$ serves
as a surrogate for $\mathbb{E}_q[\log p_\theta(z_1\mid\cdot)]$,
$\lambda_{\mathrm{VI}}D_{\mathrm{KL}}(q_\phi\|p_\psi)$ is the residual KL term, and
$\beta D_{\mathrm{KL}}(\mathrm{sg}[q_\omega]\|p_\eta)$ in $\mathcal{L}_{\mathrm{proto}}$
(Eq.~\ref{eq:prototype-objective}) is the prototype KL term, whose cross-entropy terms anchor $q_\omega$ and $p_\eta$ to
annotation-derived states; the remaining losses are auxiliary regularizers.
%%%%

Appendix~\ref{app:training} details the stage-wise objectives, loss definitions,
negative sampling, curriculum, and optimization; Algorithm~\ref{alg:facet-training}
summarizes training.

\vspace{-.5em}
\paragraph{Inference procedure.}
Figure~\ref{fig:facet-model} and Algorithm~\ref{alg:facet-inference} summarize
inference. Given sequence $S$ and context $C$, \FACET{} computes $h_S$, $z_C$,
the soft memory readout $m_S$, and the derived semantic state
$z_S=m_S+\delta z_S$. Starting from a noise latent, the pretrained,
LoRA-adapted $\varepsilon$-predictor receives the projected transport state
with $E_C(C)$, retains the direct $h_S$ pathway, and uses a separate
$m_S$-conditioned $\varepsilon$-adapter to produce the base velocity $v_L$.
The context-first hierarchical head predicts a gated correction, yielding
$v_\Theta=v_L+g(\tau)\Delta v$. The resulting SDE is integrated with
Euler-Maruyama~\citep{albergo2023stochastic,song2021sde}; the final step
yields the base clean-latent estimate $\hat z_1^L$, decoded by the frozen VAE
decoder. Training-only posterior and target-dependent reconstruction paths are
omitted. See Appendices~\ref{app:setup} and~\ref{app:flow} for details.

\vspace{-.3em}
\section{Experiments and Results}
\label{sec:results}
\vspace{-.2em}
\subsection{Experimental setting}
\label{sec:experimental-setting}
\vspace{-.4em}
\paragraph{Datasets and evaluation regimes.}
We evaluate \FACET{} on protein-held-out HPA and OpenCell benchmarks.
HPA contains 696 test images from 97 held-out proteins and provides nucleus,
endoplasmic-reticulum, and microtubule context channels; we evaluate four
compositions of these channels. OpenCell uses nucleus-only context and evaluates
both {HPA-adapted} \FACET{}, which transfers the HPA-trained
\FACET{} modules, and {OpenCell-adapted} \FACET{}, which fits
these modules on OpenCell data. Dataset construction and preprocessing are
detailed in Appendix~\ref{app:setup}.

\paragraph{Metrics and statistics.}
%
%%%%%%%%%%%%%%%%%%%%%
\begin{wrapfigure}{r}{0.65\linewidth}
    \centering
    \vspace{-.5em}
    \includegraphics[width=\linewidth]{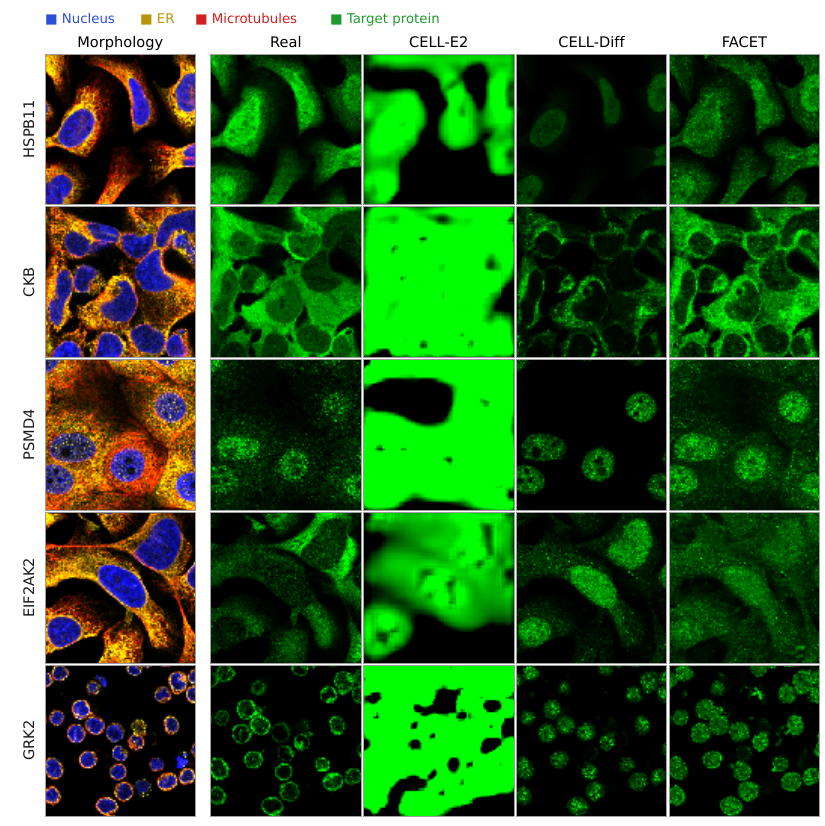}
    \vspace{-2em}
    \caption{
    Representative held-out HPA examples: morphology context, target, and model outputs. CELL-E2 uses nucleus-only context; CELL-Diff and FACET use all three context channels.
    % Representative held-out HPA examples. The context column provides the cellular-context input, while the remaining columns show the target protein image and outputs from CELL-E2, CELL-Diff, and FACET. CELL-E2 uses nucleus-only context; CELL-Diff and FACET use the full three-channel context.
    }
    \label{fig:facet_qualitative_main5}
    \vspace{-1em}
\end{wrapfigure}
%%%%%%%%%%%%%%%%%%%%%
 
%
We report MI, MS-SSIM, Haralick texture agreement, PSNR,
continuous spatial overlap ($\mathrm{IoU}_{\mathrm{cont}}$), FID, and network
function evaluations (NFE). $\mathrm{IoU}_{\mathrm{cont}}$ is the generalized Jaccard (Ruzicka) similarity. For protein-association pattern recovery,
we report Pearson-similarity separation and Mann-Whitney $U$ tests; for
predictive calibration, we report ECE, MCE, and Brier score. Metric definitions,
aggregation, statistical tests, and sampling and calibration protocols are
provided in Appendices~\ref{app:metrics} and~\ref{app:statistical-protocol}.

\vspace{-.8em}
\paragraph{Baselines and controls.}
We compare against two published generators, CELL-E2~\citep{khwaja2023celle2} and
CELL-Diff~\citep{zheng2025celldiff}, as well as several backbone-matched internal baselines: $\varepsilon$-adaptation-only control, pathway and inference-time memory ablations,
adaptation and initialization variants, and sampler variants across NFE budgets. 
To ensure that performance differences reflect FACET's conditioning design rather than
differences in backbone capacity or pretraining, all backbone-matched systems (CELL-Diff,
\FACET{}, and its internal variants) share the same pretrained $\varepsilon$-predictor, VAE,
and ESM2 protein language model from the corresponding CELL-Diff checkpoint. 
CELL-Diff is, to our knowledge, the only public sequence-conditioned latent diffusion model with
checkpoints for both benchmarks (Appendix~\ref{app:backbone-matched}).
In all \FACET{} variants, the VAE and ESM2 remain frozen; the $\varepsilon$-predictor is
either kept frozen or, in variants with backbone LoRA, adapted only through the introduced
LoRA updates.

\vspace{-.5em}
\subsection{Conditional phenotype generation on HPA}
\label{sec:hpa-results}
%
%%%%
\begin{wraptable}{r}{0.75\columnwidth}
    \centering
    \vspace{-2em}
    \caption{Comparison on the HPA test set. Higher is better for all metrics except NFE and FID. Best results are shown in bold.}
    \label{tab:hpa_main_comparison}
    \resizebox{\linewidth}{!}{%
    \begin{tabular}{lccccccc}
        \toprule
        Method
        & NFE $\downarrow$
        & MI $\uparrow$
        & MS-SSIM $\uparrow$
        & Haralick $\uparrow$
        & PSNR $\uparrow$
        & $\mathrm{IoU}_{\mathrm{cont}}$ $\uparrow$
        & FID $\downarrow$\\
        \midrule
        CELL-E2   & --           & 0.4041          & 0.1370          & 0.5644          & 6.36           & 0.1755          & 166.4\\
        CELL-Diff & 100          & 0.5355          & 0.3471          & 0.5182          & 15.77          & 0.2478          & 45.60\\
        FACET     & \textbf{25}  & \textbf{0.5481} & \textbf{0.3651} & \textbf{0.5854} & \textbf{16.15} & \textbf{0.3328} & \textbf{33.20}\\
        \bottomrule
    \end{tabular}%
    }
    \vspace{-1em}
\end{wraptable}
%%%%
%
% %
% %%%%
% \begin{wraptable}{r}{0.72\columnwidth}
%     \centering
%     \vspace{-2.5em}
%     \caption{Comparison on the HPA test set. Higher is better for all metrics except NFE and FID. Best results are shown in bold.}
%     \label{tab:hpa_main_comparison}
%     \resizebox{\linewidth}{!}{%
%     \begin{tabular}{lccccccc}
%         \toprule
%         Method
%         & NFE $\downarrow$
%         & MI $\uparrow$
%         & MS-SSIM $\uparrow$
%         & Haralick $\uparrow$
%         & PSNR $\uparrow$
%         & $\mathrm{IoU}_{\mathrm{cont}}$ $\uparrow$
%         & FID $\downarrow$\\
%         \midrule
%         CELL-E2   & --           & 0.4041          & 0.1370          & 0.5644          & 6.36           & 0.1755          & 166.4\\
%         CELL-Diff & 100          & 0.5355          & 0.3471          & 0.5182          & 15.77          & 0.2478          & 45.6\\
%         FACET     & \textbf{25}  & \textbf{0.5481} & \textbf{0.3651} & \textbf{0.5854} & \textbf{16.15} & \textbf{0.3328} & \textbf{33.20}\\
%         \bottomrule
%     \end{tabular}%
%     }
%     \vspace{-1.5em}
% \end{wraptable}
% %%%%

%
%%%
\begin{table}[b]
    \centering
    \vspace{-1.8em}
    \caption{Performance across cellular-context compositions on HPA. All systems use the same
    proteins, sequence inputs, and sampling protocol per composition. Bold marks the better value
    within each composition.}
    \label{tab:context-composition}
    \small
    \setlength{\tabcolsep}{5pt}
    \resizebox{0.85\linewidth}{!}{%
    \begin{tabular}{@{}llcccccc@{}}
    \toprule
    Context & System & MI $\uparrow$ & MS-SSIM $\uparrow$ & Haralick $\uparrow$ & PSNR $\uparrow$
    & $\mathrm{IoU}_{\mathrm{cont}}$ $\uparrow$ & FID $\downarrow$ \\
    \midrule
    \multirow{2}{*}{Nucleus}
    & CELL-Diff & \textbf{0.3091} & \textbf{0.2777} & 0.4932 & \textbf{15.32} & 0.2047 & \textbf{51.10} \\
    & \FACET{}  & 0.2519 & 0.1712 & \textbf{0.5010} & 14.66 & \textbf{0.2768} & 52.12 \\
    \addlinespace[2pt]
    \multirow{2}{*}{Nucleus + ER}
    & CELL-Diff & 0.4562 & 0.3154 & 0.4898 & 15.52 & 0.2182 & 60.00 \\
    & \FACET{}  & \textbf{0.5120} & \textbf{0.3413} & \textbf{0.5726} & \textbf{15.96} & \textbf{0.3268} & \textbf{33.83} \\
    \addlinespace[2pt]
    \multirow{2}{*}{Nucleus + MT}
    & CELL-Diff & 0.5097 & 0.3404 & 0.5196 & 15.60 & 0.2435 & 47.60 \\
    & \FACET{}  & \textbf{0.5226} & \textbf{0.3524} & \textbf{0.5796} & \textbf{16.03} & \textbf{0.3266} & \textbf{33.24} \\
    \addlinespace[2pt]
    \multirow{2}{*}{Nucleus + ER + MT}
    & CELL-Diff & 0.5355 & 0.3471 & 0.5182 & 15.77 & 0.2478 & 45.60 \\
    & \FACET{}  & \textbf{0.5481} & \textbf{0.3651} & \textbf{0.5854} & \textbf{16.15} & \textbf{0.3328} & \textbf{33.20} \\
    \bottomrule
    \end{tabular}%
    }
    \vspace{-0.1em}
\end{table}
%%%
%

\vspace{-.5em}
\subsubsection{Results on HPA}
\vspace{-.5em}
We first evaluate protein-conditioned spatial-phenotype generation on the held-out HPA test set. 
Tab.~\ref{tab:hpa_main_comparison} summarizes the comparison across image-quality metrics and sampling cost. \FACET{} achieves the best value on all six reported image-quality metrics while requiring only 25 network function evaluations per image, compared with 100 for CELL-Diff. 
% Relative to the baseline, it improves continuous intensity overlap from $0.2478$ to $0.3328$ and lowers FID from $45.6$ to $33.20$, while using 75\% fewer network evaluations. The remaining image-quality metrics show the same direction of improvement. 
Relative to the baseline, it improves continuous intensity overlap from 0.2478 to 0.3328 and
lowers FID from 45.6 to 33.20. Despite its added modules, \FACET{} also reduces
generation time for the 696-image test set from 78 to 15 minutes on the same GPU
(Appendix~\ref{app:sampling-efficiency}).
Fig.~\ref{fig:facet_qualitative_main5} and Appendix Figs.~\ref{fig:app-hpa-same-protein},~\ref{fig:app-hpa-grid}, and~\ref{fig:app-hpa-comparison} show representative qualitative results.

\vspace{-.6em}
\subsubsection{Performance across cellular-context compositions}
\vspace{-.1em}
\vspace{-.5em}
Tab.~\ref{tab:context-composition} evaluates the same checkpoint across four cellular-context compositions.
\FACET{} improves every reported metric in the two partial multi-channel regimes and the full-context regime. 
A withheld morphology channel is replaced by the pretrained predictor's learned
placeholder embedding, in both the context canvas $E_C(C)$ and the features
from which $z_C$ is computed, so every system sees the same ablated context.
Under nucleus-only context, it retains gains in continuous intensity overlap and Haralick agreement, whereas CELL-Diff is higher on the remaining reported metrics. Representative qualitative results are depicted in Appendix Fig.~\ref{fig:app-context-ablation}.

% \paragraph{Qualitative results.} Fig.~\ref{fig:facet_qualitative_main5}, Appendix Fig.~\ref{fig:app-opencell-comparison} and Appendix Fig.~\ref{fig:app-context-ablation} provide representative qualitative evidence across proteins and cellular-context compositions.

\vspace{-.6em}
\subsection{Cross-dataset {evaluation}}
\label{sec:opencell-results}
\vspace{-.2em}
%%%%%%%%%
\begin{wraptable}{r}{0.81\columnwidth}
    \vspace{-2.2em}
    % \caption{Comparison on the OpenCell test set. Higher is better for all
    % metrics except NFE and FID. Best results are shown in bold. Only the
    % nucleus images are used as the cellular context. {The HPA-adapted row transfers FACET-specific modules from HPA, whereas the OpenCell-adapted row fits them on OpenCell training data.}}
    \caption{Comparison on the OpenCell test set (nucleus-only context). All systems except CELL-E2 share the OpenCell-trained CELL-Diff backbone; HPA-adapted transfers the \FACET{} modules from HPA without refitting, whereas OpenCell-adapted fits them on OpenCell. Best results are in bold.}
    \label{tab:opencell-main-comparison}
    \centering
    \resizebox{\linewidth}{!}{%
    \begin{tabular}{lccccccc}
    \toprule
    \textbf{Method}
    & {NFE $\downarrow$}
    & {MI $\uparrow$}
    & {MS-SSIM $\uparrow$}
    & {Haralick $\uparrow$}
    & {PSNR $\uparrow$}
    & {IoU$_{\mathrm{cont}}$ $\uparrow$}
    & {FID $\downarrow$}\\
    \midrule
    
    CELL-E2 & -- & 0.1649 & 0.1020 & 0.4542 & 11.31 & 0.1779 & {245.8}\\
    
    CELL-Diff
    & 100
    & 0.2732
    & \textbf{0.4078}
    & 0.3663
    & 13.20
    & 0.3722
    & {68.90}\\
    
    FACET (HPA-adapted)
    & \textbf{25}
    & 0.2753
    & 0.3653
    & 0.5177
    & \textbf{13.85}
    & 0.4222
    & 43.33\\
    
    FACET (OpenCell-adapted)
    & \textbf{25}
    & \textbf{0.2998}
    & 0.3893
    & \textbf{0.5199}
    & 13.75
    & \textbf{0.4244}
    & \textbf{36.83}\\
    \bottomrule
    \end{tabular}%
    }
    \vspace{-1em}
\end{wraptable}
%%%%%%%%%
% \subsubsection{{\revnew{remove subsubsection title, keep original}HPA-module transfer and OpenCell adaptation}}
We next evaluate HPA-module transfer on OpenCell, using the nucleus image as the cellular context. 
For both CELL-Diff and FACET, the pretrained $\varepsilon$-predictor backbone, VAE encoder/decoder, and ESM2 protein language model are taken from the OpenCell CELL-Diff checkpoint. The HPA-adapted FACET variant transfers the FACET-specific modules from HPA, whereas the
OpenCell-adapted variant fit these modules directly on OpenCell training data. 
As shown in Tab.~\ref{tab:opencell-main-comparison}, HPA-module transfer improves MI slightly, Haralick agreement, PSNR, continuous intensity overlap, and FID relative to CELL-Diff, although MS-SSIM is lower. OpenCell-adaptation improves MI, MS-SSIM, Haralick agreement, continuous intensity overlap, and FID relative to the HPA-adapted variant, with only a small decrease in PSNR. See qualitative results in Appendix Fig.~\ref{fig:app-opencell-comparison}.

\vspace{-.6em}
\subsection{Spatially resolved protein-association pattern recovery} \label{sec:protein-association-recovery}
\vspace{-.2em}
%%%%
\begin{wraptable}{r}{0.60\columnwidth}
\centering
\vspace{-2em}
\caption{Spatially resolved protein-association pattern recovery under a
localization-matched evaluation adapted from \citep{proticelli}. The baseline denotes CELL-Diff. Pearson
similarity is computed after removing the per-cell mean image. $\Delta$
measures the separation between STRING-associated and matched non-associated
pairs; positive values indicate stronger recovery.}
\label{tab:protein-association-recovery}
\small
\setlength{\tabcolsep}{5pt}
\renewcommand{\arraystretch}{1.08}

\resizebox{\linewidth}{!}{%
\begin{tabular}{@{}lcccc@{}}
\toprule
System
& \shortstack{STRING score\\threshold}
& \shortstack{Pairs\\(assoc./null)}
& Pearson $\Delta$ $\uparrow$
& $p_{\mathrm{assoc.\ vs.\ null}}$ \\
\midrule
Baseline
& 0.85
& 47/2659
& +0.1005
& 0.298 \\

FACET
& 0.85
& 47/2659
& \textbf{+0.1875}
& $1.3\times10^{-3}$ \\

\hline
Baseline
& 0.70
& 65/2641
& +0.0954
& 0.170 \\

FACET
& 0.70
& 65/2641
& \textbf{+0.1718}
& $4.5\times10^{-4}$ \\

\hline
Baseline
& 0.50
& 112/2594
& +0.0922
& 0.042 \\

FACET
& 0.50
& 112/2594
& \textbf{+0.1369}
& $2.5\times10^{-4}$ \\

\hline
Baseline
& 0.40
& 164/2542
& +0.0925
& 0.0093 \\

FACET
& 0.40
& 164/2542
& \textbf{+0.1289}
& $1.7\times10^{-5}$ \\
\bottomrule
\end{tabular}%
}
\vspace{-1em}
\end{wraptable}
%%%%
To assess whether generated images preserve protein-association patterns, we
adapt the localization-matched evaluation of \cite{proticelli}. We use
STRING~\citep{szklarczyk2023string}, a database of known and predicted
functional protein associations, solely to obtain external evaluation labels.
We generate the 97 proteins in the HPA test split under eight fixed cellular contexts, using matched context and noise seeds across systems. For each protein pair, we compute within-cell Pearson similarity after removing the per-cell mean image across proteins. This suppresses similarity caused by shared cellular morphology. We then compare STRING-associated pairs with non-associated pairs that share at least one annotated compartment. STRING labels are used only for evaluation and do not enter model training or conditioning.

We report the difference
\[
\Delta =
\mathbb{E}\!\left[r(a,b)\mid\mathrm{STRING\text{-}associated}\right]
-
\mathbb{E}\!\left[r(a,b)\mid\mathrm{matched\ null}\right],
\]
where positive values indicate that STRING-associated proteins are rendered
more similarly. 
As shown in Tab.~\ref{tab:protein-association-recovery},
\FACET{} produces a larger positive separation than CELL-Diff at all four
STRING thresholds. At STRING score $\geq 0.85$, \FACET{} increases $\Delta$
from $+0.1005$ to $+0.1875$, while at STRING score $\geq 0.40$, it increases
$\Delta$ from $+0.0925$ to $+0.1289$. All four \FACET{} contrasts are
significant under the two-sided Mann-Whitney $U$ test~\citep{mann1947test}, with
$p$ values ranging from $1.7\times10^{-5}$ to $1.3\times10^{-3}$. These
results indicate that \FACET{} recovers association-consistent spatial
patterns beyond shared cellular context.
% \textcolor{orange}{where positive values indicate that STRING-associated proteins are rendered more similarly. As shown in Tab.~\ref{tab:protein-association-recovery}, \FACET{} produces a larger positive separation than CELL-Diff at all four STRING thresholds. At STRING score $\geq 0.85$, \FACET{} increases $\Delta$ from $+0.1005$ to $+0.1875$, while at STRING score $\geq 0.40$, it increases $\Delta$ from $+0.0925$ to $+0.1289$. The reported $p$-values assess associated-versus-matched-null comparisons within each system; they do not test whether the difference between \FACET{} and CELL-Diff is statistically significant. We therefore interpret the consistent between-system increase descriptively as evidence that \FACET{} recovers association-consistent spatial patterns beyond shared cellular context.}

\vspace{-.6em}
\subsection{Component ablations and additional analyses}
\label{sec:ablations}
\vspace{-.4em}

\subsubsection{Pathway ablation}
\vspace{-.4em}
%%%%
\begin{wrapfigure}{r}{0.5\linewidth}
    \centering
    \vspace{-2em}
    \includegraphics[width=\linewidth]{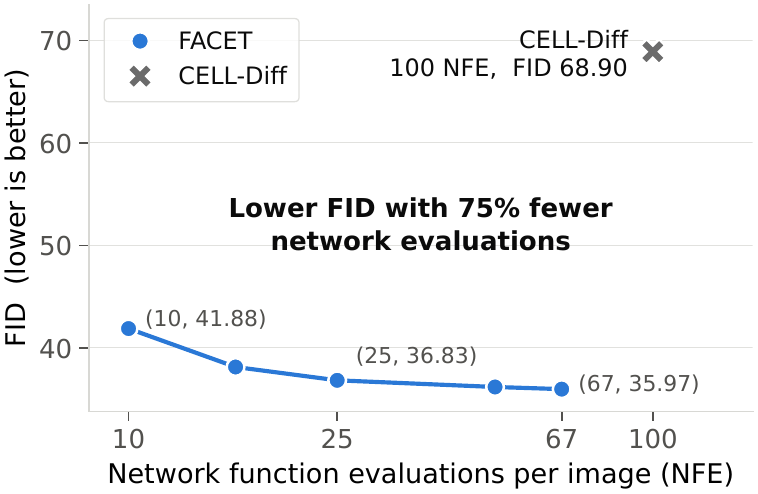}
    \vspace{-2em}
    \caption{Quality–cost comparison on OpenCell.}
    \label{fig:quality-cost-opencell}
    \vspace{-2em}
\end{wrapfigure}
%%%%%
We compare controlled FACET configurations under a common evaluation protocol. Tab.~\ref{tab:hpa-facet-ablation} shows that the complete configuration performs best. Adding the sequence-conditioned semantic pathway raises $\mathrm{IoU}_{\mathrm{cont}}$ from $0.2581$ to $0.2826$, while adding either the backbone LoRA or the hierarchical velocity-correction head provides further improvement. The full configuration, which includes both components, achieves the highest observed value of $0.3328$.

\begin{table}[!t]
\centering
\vspace{-1.2em}
\caption{Controlled FACET configuration ablation on the HPA benchmark.
Each configuration is evaluated on the same held-out proteins using identical
cellular context, sampler, and preprocessing.
Checkmarks indicate active pathways. ``hier.'' represents ``hierarchical''.}
\label{tab:hpa-facet-ablation}
\small
\setlength{\tabcolsep}{3.5pt}
\renewcommand{\arraystretch}{1.1}

\resizebox{\linewidth}{!}{%
\begin{tabular}{@{}p{0.36\linewidth}ccccc@{}}
\toprule
\shortstack[l]{FACET configuration}
&
\shortstack{$\varepsilon$-adapter}
&
\shortstack{Memory-conditioned\\semantic pathway}
&
\shortstack{Backbone LoRA\\(base velocity $v_L$)}
&
{\shortstack{Hierarchical head\\(velocity correction $\Delta v$)}}
&
\shortstack{$\mathrm{IoU}_{\mathrm{cont}}$$\uparrow$}
\\
\midrule

FACET (full)
& \checkmark
& \checkmark
& \checkmark
& \checkmark
& \textbf{0.3328}
\\

Memory-conditioned + LoRA
& \checkmark
& \checkmark
& \checkmark
& --
& 0.3236
\\

Memory-conditioned + hier. head
& \checkmark
& \checkmark
& --
& \checkmark
& 0.3052
\\

Memory-conditioned
& \checkmark
& \checkmark
& --
& --
& 0.2826
\\

$\varepsilon$-adaptation-only
& \checkmark
& --
& --
& --
& 0.2581
\\

\bottomrule
\end{tabular}%
}
\vspace{-2em}
% \parbox{\linewidth}{\footnotesize
% \textit{Note.}
% The sequence-conditioned semantic pathway includes the prototype memory,
% paired sequence objective, and fine protein-specific residual. Backbone LoRA
% modifies the base velocity $v_L$, while the hierarchical head produces the
% correction $\Delta v$ in $v=v_L+g(\tau) \Delta v$. In the head-only configuration, the
% backbone LoRA is held at zero and is therefore shown as inactive.
% }
\end{table}

\vspace{-.6em}
\subsubsection{Quality-cost trade-off}
\vspace{-.4em}
% %%%%
% \begin{wrapfigure}{r}{0.5\linewidth}
%     \centering
%     \vspace{-2em}
%     \includegraphics[width=\linewidth]{figures/facet_quality_cost.pdf}
%     \vspace{-2em}
%     \caption{Quality–cost comparison on OpenCell.}
%     \label{fig:quality-cost-opencell}
%     \vspace{-1em}
% \end{wrapfigure}
% %%%%%
The quality-cost comparison is summarized in Fig.~\ref{fig:quality-cost-opencell}. The OpenCell-adapted \FACET{} model reaches lower FID with substantially fewer network function evaluations than CELL-Diff. At 25 NFE, it achieves FID $=36.83$, compared with $68.90$ for CELL-Diff at 100 NFE, corresponding to a 75\% reduction in network evaluations per image.

% \vspace{-.7em}
\subsubsection{Ablation on adaptation configurations}
\vspace{-.4em}
To examine the roles of initialization and OpenCell adaptation, we compare the corresponding configurations in Tab.~\ref{tab:opencell_adaptation_ablation}. The comparison shows that performance is sensitive to both the initialization of the FACET-specific modules and whether the semantic memory is refitted on OpenCell data. The cold-start configuration with OpenCell-fitted memory gives the highest overlap ($0.4244$), while refitting the memory in the warm-start setting nearly recovers the cold-start performance.

\begin{table}[!t]
    \centering
    \vspace{.5em}
    \caption{{HPA-module transfer and OpenCell adaptation ablation.} A checkmark indicates that the corresponding component is active. Note that, for all the configurations, the {$\varepsilon$-predictor backbone, VAE encoder/decoder, and protein LM} are adopted from the OpenCell checkpoint by \cite{zheng2025celldiff}.}
    \label{tab:opencell_adaptation_ablation}
    \resizebox{\linewidth}{!}{%
    \begin{tabular}{lcccc}
        \toprule
        {Configuration}
        & {OpenCell adaptation}
        & {HPA module initialization}
        & {OpenCell-fit memory}
        & \textbf{$\mathrm{IoU}_{\mathrm{cont}}$} $\uparrow$ \\
        \midrule

        FACET, HPA-adapted
        & --
        & $\checkmark$
        & --
        & 0.4222 \\

        FACET, OpenCell-adapted, cold start
        & $\checkmark$
        & --
        & $\checkmark$
        & \textbf{0.4244} \\

        Warm start, inherited HPA memory
        & $\checkmark$
        & $\checkmark$
        & --
        & 0.4155 \\

        Warm start, refitted OpenCell memory
        & $\checkmark$
        & $\checkmark$
        & $\checkmark$
        & 0.4233 \\

        \bottomrule
    \end{tabular}%
    }
    \vspace{-1.5em}
\end{table}

% \vspace{-.5em}
\subsubsection{Ablation of inference-time memory}
\vspace{-.5em}
To isolate the contribution of the learned semantic memory at inference, we
keep all other FACET components fixed and replace the learned memory with
either the uniform mean of the prototypes or a zero memory. As shown in
Tab.~\ref{tab:hpa_memory_ablation}, the unmodified learned memory achieves
the highest $\mathrm{IoU}_{\mathrm{cont}}$ of $0.3328$. Replacing it with the
uniform prototype mean or zero memory reduces performance to $0.2820$ and
$0.3014$, respectively. 
Zero memory degrades less because the standardized PCA space is centered, so zero approximates the frequency-weighted prototype average, whereas the uniform mean overweights rare localization categories and injects a protein-agnostic bias.
This result indicates that FACET benefits from the
protein-conditioned organization of the learned memory rather than from memory
injection alone.

\begin{table}[!t]
    \centering
    \caption{Ablation of inference-time memory on HPA. A checkmark indicates that the corresponding memory configuration is active.}
    \label{tab:hpa_memory_ablation}
    \resizebox{\linewidth}{!}{%
    \begin{tabular}{lccc}
        \toprule
        {Configuration}
        & {Learned memory}
        & {Memory replaced at inference}
        & {$\mathrm{IoU}_{\mathrm{cont}}$} $\uparrow$ \\
        \midrule

        FACET, unmodified memory
        & $\checkmark$
        & --
        & \textbf{0.3328} \\

        FACET, uniform prototype mean
        & --
        & $\checkmark$
        & 0.2820 \\

        FACET, zero memory
        & --
        & $\checkmark$
        & 0.3014 \\

        \bottomrule
    \end{tabular}%
    }
    \vspace{-1.5em}
\end{table}

\vspace{-.6em}
\subsubsection{Sampler ablation}
\vspace{-.5em}
%%%
\begin{wraptable}{r}{0.40\columnwidth}
\centering
\vspace{-4em}
\caption{OpenCell sampling results across solvers and network function evaluations (NFE).}
\label{tab:opencell-samplers}
\small
\setlength{\tabcolsep}{4pt}
\resizebox{\linewidth}{!}{%
\begin{tabular}{@{}lccc@{}}
\toprule
Sampler & NFE & FID $\downarrow$ & $\mathrm{IoU}_{\mathrm{cont}}$ $\uparrow$ \\
\midrule
SDE & 10 & 41.88 & 0.4301 \\
% SDE & 16 & 38.13 & 0.4263 \\
SDE & 25 & 36.83 & 0.4244 \\
SDE & 50 & 36.18 & 0.4214 \\
% SDE & 67 & 35.97 & 0.4200 \\
\addlinespace
ODE (Heun's) & 25 & 36.60 & 0.4205 \\
ODE (Heun's) & 50 & 36.17 & 0.4187 \\
\bottomrule
\end{tabular}%
}
\vspace{-2em}
\end{wraptable}
%%%
We compare the Euler-Maruyama SDE sampler with a deterministic probability-flow ODE integrated
using Heun's method \citep{karras2022elucidating} on OpenCell (Tab.~\ref{tab:opencell-samplers}).
At matched NFE, both samplers reach nearly identical FID, while the SDE sampler yields slightly
higher $\mathrm{IoU}_{\mathrm{cont}}$. Both remain well below CELL-Diff at 100 NFE (FID 68.90),
indicating that FACET's efficiency gains are not specific to the choice of solver.
% We compare the Euler--Maruyama SDE sampler with a deterministic probability-flow ODE integrated
% using Heun's method \citep{karras2022elucidating} on OpenCell (Tab.~\ref{tab:opencell-samplers}).
% At matched NFE, both samplers reach nearly identical FID (36.83 vs.\ 36.60 at 25 NFE; 36.18 vs.\
% 36.17 at 50 NFE), while the SDE sampler yields slightly higher $\mathrm{IoU}_{\mathrm{cont}}$ at
% both budgets. For the SDE sampler, FID improves from 41.88 to 36.83 between 10 and 25 NFE and
% changes by less than 0.9 beyond that, whereas $\mathrm{IoU}_{\mathrm{cont}}$ varies by at most
% 0.01 across budgets. Every configuration remains well below CELL-Diff at 100 NFE (FID 68.90),
% indicating that FACET's efficiency gains are not specific to the choice of solver and
% motivating 25 NFE as the default operating point.

\subsubsection{Predictive calibration}
%%%%
\begin{wraptable}{r}{0.55\columnwidth}
\centering\small\setlength{\tabcolsep}{10pt}
\vspace{-2.8em}
\caption{{Predictive calibration on HPA.} One temperature per model, fit on a
protein-disjoint validation split and held fixed on test. Foreground probabilities
come from 16 predictive samples per input. }
\label{tab:res-calibration}
\begin{tabular}{@{}lrrr@{}}
\toprule
Method & ECE~$\downarrow$ & MCE~$\downarrow$ & Brier~$\downarrow$\\
\midrule
{CELL-Diff} & 0.21501 & 0.24886 & 0.23372\\
\FACET{} & \textbf{0.05442} & \textbf{0.14850} & \textbf{0.19915}\\
\bottomrule
\end{tabular}
\vspace{-.5em}
\end{wraptable}
%%%%
To evaluate calibration and overall probabilistic forecast quality, we report ECE, MCE, and the Brier score using 16 samples per input. We follow the post-hoc temperature-scaling procedure of \cite{guo2017calibration}, fitting the temperature on a protein-disjoint validation partition and keeping it fixed for test evaluation.
As shown in Tab.~\ref{tab:res-calibration}, \FACET{}
substantially improves all three calibration measures: ECE decreases from
$0.21501$ to $0.05442$, MCE from $0.24886$ to $0.14850$, and Brier score from
$0.23372$ to $0.19915$. These results indicate that \FACET{} produces more
reliable predictive foreground probabilities. This provides additional
evidence that the factorized stochastic predictor is not only accurate but
also better calibrated under repeated sampling.

% \subsection{Summary of experimental findings}
% \label{sec:results-summary}
% Across the experiments, \FACET{} improves spatial and distributional fidelity on HPA while requiring 75\% less network function evaluations per image. Its HPA-trained FACET modules retain important gains under OpenCell evaluation, and further OpenCell adaptation improves most reported metrics. The localization-matched analysis shows a larger descriptive separation between STRING-associated and matched non-associated protein pairs, indicating recovery of association-consistent spatial patterns beyond shared cellular context. Controlled configuration comparisons are consistent with the roles of the semantic pathway, backbone LoRA, and hierarchical velocity correction, while the calibration analysis shows lower post-hoc calibrated forecast errors. Together, these findings are consistent with \FACET{}'s central design principle: organizing sequence semantics, cellular context, and residual variation into role-specific conditioning pathways enables stronger conditional generation with efficient stochastic transport.

% \subsection{Limitations and scope} FACET builds on a pretrained variance-preserving backbone
% and inherits its finite schedule, which supports up to 67 NFE (Appendix~\ref{app:flow}).
% Extending the framework to 3D and time-lapse imaging, joint generation of multiple proteins
% per cell, and additional conditions such as perturbations are natural next steps.
\vspace{-.2em}
\section{Conclusion}
\vspace{-.2em}
We introduced \FACET{}, a factorized probabilistic framework that turns the asymmetric,
hierarchical structure of protein localization into an explicit inductive bias atop a pretrained
diffusion prior. On unseen proteins across two benchmarks, it substantially improves spatial
and distributional fidelity at a fraction of the sampling cost, while improving
protein-association recovery and calibration.
With the backbone held fixed, these gains are attributable to \FACET{}'s factorized
probabilistic formulation rather than generator capacity, making it complementary to stronger
backbones and a template for conditional generation whenever inputs differ in spatial alignment
and explanatory power.

\vspace{-.5em}
\paragraph{Scope and future directions.} \FACET{} inherits its backbone's finite noise schedule,
which bounds the usable sampling steps; continuous-time variance-preserving backbones lift this
directly. Extending \FACET{} to 3D and time-lapse imaging, multi-protein generation, and
perturbations are natural next steps toward in silico spatial proteomics.

\bibliography{ref}
\bibliographystyle{iclr2027_conference}

\appendix

\section{Related Work}
\label{app:related}

\paragraph{Asymmetric multimodal conditioning.}
Joint multimodal training can under-utilize one modality even when all inputs are
informative. Gradient Blending identified different optimization and
generalization rates across modalities \citep{wang2020multimodal}, gradient
modulation and prototype rebalancing mitigate modality imbalance
\citep{peng2022ogm,fan2023pmr}, and theoretical work shows that joint gradient
descent can produce modality competition in simplified settings
\citep{huang2022competition}. These works primarily study discriminative fusion.
{\FACET{} addresses a distinct generative setting in which conditions
differ in representation, spatial alignment, and explanatory role, assigning
them distinct but coupled generative roles rather than rebalancing their
contributions to a shared discriminative objective.}

\paragraph{Protein and cellular image generation.}
{Protein-localization image generation has been approached with
bidirectional sequence--image transformers conditioned on nucleus context
\citep{khwaja2023celle2}, sequence- and landmark-conditioned image prediction
\citep{zhang2025prediction}, diffusion conditioned on protein sequence and cellular
morphology \citep{zheng2025celldiff}, conditional diffusion using cellular
landmarks and learned protein identities \citep{proticelli}, and recent
channel-adaptive autoregressive generation from sequence and cellular landmarks
\citep{kambhampati2026vermeer}.}
Perturbation-conditioned diffusion models instead predict morphology from
transcriptomic or perturbation information \citep{wang2025morphdiff}. HPA
\citep{thul2017hpa} and OpenCell \citep{cho2022opencell} provide complementary
protein-localization microscopy resources.
{Our focus is not a new image backbone, but how sequence-dependent
localization and cellular morphology are structured within a probabilistic
conditional generator, including an architecture-controlled comparison to
CELL-Diff.}

\paragraph{Diffusion, stochastic flows, and few-step transport.}
Diffusion models construct generative processes through progressive corruption
and denoising, latent diffusion
moves this process to a compressed image representation
\citep{rombach2022ldm,ho2020ddpm,song2021ddim,song2021sde,mahbub2025uncertainty,ellington2024accurate,sun2024mixture,zou2024large}. Flow matching learns continuous vector fields from
tractable conditional paths \citep{lipman2023flow}, rectified flow emphasizes
direct transport paths \citep{liu2023rectified}, stochastic interpolants unify
deterministic and stochastic transports \citep{albergo2023stochastic}, and
optimal-transport couplings can improve flow training \citep{tong2024otcfm}.
{Recent methods bridge pretrained diffusion and flow models through
interpolant and timestep alignment \citep{schusterbauer2025diff2flow}, or
accelerate few-step generation by learning stochastic-interpolant schedules
\citep{min2026bezierflow}. \FACET{} uses stochastic flow matching as its
transport component and introduces an explicit variance-preserving state
projection that makes continuous transport states compatible with a pretrained
finite-schedule diffusion predictor.}

\paragraph{Hierarchical latent variables and parameter-efficient conditioning.}
Variational autoencoders provide amortized latent-variable inference, while hierarchical variants show how multiple stochastic
levels can represent different factors of variation \citep{kingma2014vae,sonderby2016ladder,mahbub2026retrieval,mahbub2026prism}.
Feature-wise Linear Modulation (FiLM) provides a general mechanism for
conditional affine modulation \citep{perez2018film}. Low-Rank Adaptation (LoRA)
trains additive low-rank weight updates while retaining pretrained parameters
\citep{hu2022lora}, and ControlNet demonstrates parameter-efficient conditioning
around a strong pretrained image prior \citep{zhang2023controlnet}.
{\FACET{} uses variational residual inference, FiLM, and LoRA within an
asymmetric factorization that assigns cellular context, coarse sequence
semantics, fine protein-specific variation, and image stochasticity distinct
but coupled generative roles.}

\paragraph{Context residualization, shortcuts, and invariance.}
Shortcut learning concerns reliance on easy predictive cues
\citep{geirhos2020shortcut}, batch effects arise from acquisition or processing
variation \citep{leek2010batch}, and domain-adversarial learning suppresses
domain-identifying information to obtain invariance \citep{ganin2016dann}.
{\FACET{} differs from these objectives: cellular morphology is a
scientifically meaningful conditioning variable and remains directly available
to the generator rather than being treated as a nuisance signal to remove.
Context residualization is applied only when constructing the semantic memory,
encouraging sequence semantics to capture localization information beyond what
context already explains rather than enforcing invariance to context. The
objective remains high-fidelity conditional generation from both sequence and
cellular context, not removal of either condition.}

\section{Additional Method Details}
\label{app:method-details}

\subsection{Training objective and curriculum}
\label{app:training}

The probabilistic formulation in the main text specifies the roles of the
prototype-valued state $R$, continuous residual $\delta z_S$, and image
latent $z_1$. Rather than optimizing the ELBO jointly, we adopt a modular
training procedure: we first construct the context-residualized memory and
pretrain the semantic prototype prior and $\varepsilon$-adapter, then train the
continuous residual and conditional transport model.

\paragraph{Discrete semantic objective.}
Let $r$ denote the annotation-derived prototype state for a training protein.
The discrete semantic components are fit using
\begin{equation}
\mathcal L_{\rm proto}
=\mathrm{CE}(q_\omega,r)
+\beta D_{\mathrm{KL}}\!\left(\mathrm{sg}[q_\omega]\,\|\,p_\eta\right)
+\gamma\,\mathrm{CE}(p_\eta,r),
\label{eq:prototype-objective}
\end{equation}
where $\mathrm{sg}[\cdot]$ denotes stop-gradient. The
cross-entropy term trains $q_\omega$ against the annotation-derived state;
only in the KL term is $q_\omega$ treated as a stop-gradient target for
$p_\eta$. Thus, the divergence updates the prior without moving the posterior
in the reverse direction.

\paragraph{Memory-conditioned $\varepsilon$-adapter pretraining.}
We pretrain the $\varepsilon$-adapter using noise prediction, sequence ranking,
a prior KL term $D_{\mathrm{KL}}(\operatorname{sg}[q_\omega]\|p_\eta)$,
and an adapter-output penalty, while continuing to update $(\omega,\eta)$.
The $\varepsilon$-adapter and prototype prior are then frozen for transport training.

\paragraph{Flow-matching and reconstruction losses.}
As derived in Appendix~\ref{app:base-velocity-derivation}, for a sampled $\tau$, the conditional flow target is
\begin{equation}
u_\tau^\star=z_1+\dot Q_\tau\epsilon_\tau,
\qquad
\mathcal L_{\rm FM}
=\mathbb E\left[\left\|v_\Theta-u_\tau^\star\right\|_2^2\right].
\label{eq:flow-objective}
\end{equation}

{For reconstruction supervision, we derive a \textit{training-time} auxiliary
latent estimate by retaining the adapted predictor's noise
estimate $\hat\epsilon_L$:
\begin{equation}
\hat z_1^{\,\theta}
% :=v_\Theta-\dot Q_\tau\hat\epsilon_L
:=\hat z_1^L+g(\tau)\Delta v,
\label{eq:auxiliary-clean-estimate}
\end{equation} 
which follows from the velocity
decomposition in Eqs.~\ref{eq:base-velocity} and~\ref{eq:velocity}; its derivation is provided in Appendix~\ref{app:auxiliary-clean-derivation}. This fixed-noise assignment is a training-time auxiliary construction, not the inference emission.
% The second equality follows from Eqs.~\ref{eq:base-velocity} and~\ref{eq:velocity}, and the detailed derivation is provided in Appendix~\ref{app:auxiliary-clean-derivation}.
}
Let $D$ denote the frozen decoder and $d$ the residual dimension. The auxiliary terms are
\begin{align}
\mathcal L_{\rm rec}
&=\mathbb E\left[\left\|\hat z_1^{\,\theta}-z_1\right\|_2^2\right]\label{eq:reconstruction-objectives},
&
\mathcal L_{\rm rec,pix}
&=\mathbb E\left[\left\|D(\hat z_1^{\,\theta})-D(z_1)\right\|_2^2\right],
\\
\mathcal L_{\rm shrink}
&=\mathbb E\left[\left\|g(\tau)\Delta v\right\|_2^2\right]\label{eq:regularization-objectives},
&
\mathcal L_{\rm same}
&=\mathbb E\left[
\frac{\left\|\mu_q(U_Y^{(i)},m_S)-\mu_q(U_Y^{(j)},m_S)\right\|_2^2}
{d\,s_m^2}\right].
\end{align}
Here $i$ and $j$ index two images of the same protein. The reconstruction terms supervise the auxiliary latent estimate,
while flow matching supervises the composed velocity. The shrinkage term keeps the
hierarchical branch a correction to the base estimate field. The same-protein term
encourages the residual representation to capture protein-specific rather than
cell-specific variation, consistent with the sequence-conditioned prior.

\paragraph{Sequence-ranking objective.}
For a target image, we evaluate the flow-matching loss using the matched sequence
and $n$ mismatched sequences while holding the interpolant state, timestep, and
noise draws fixed. This produces a contrastive sequence-ranking loss
\begin{equation}
\mathcal L_{\rm rank}
=-\log
\frac{\exp(-\mathcal L_{\rm FM}^{+}/\tau_c)}
{\exp(-\mathcal L_{\rm FM}^{+}/\tau_c)
 +\sum_{j=1}^{n}\exp(-\mathcal L_{\rm FM}^{(j)}/\tau_c)},
\label{eq:ranking-objective}
\end{equation}
{Here $\mathcal L_{\rm FM}^{+}$ and
$\mathcal L_{\rm FM}^{(j)}$ denote the per-candidate squared velocity errors
$\|v_\Theta-u_\tau^\star\|_2^2$, not separately averaged dataset losses.
The matched training candidate uses its target-informed residual posterior;
each mismatched candidate uses its own sequence-conditioned prior.}
Negatives are sampled by first selecting
a localization prototype other than the target's uniformly and then sampling a
training protein uniformly within that prototype. Each negative is evaluated
using its own sequence prior, matching the inference-time use of the model.

\paragraph{Implemented flow-stage objective.}
The final flow-stage objective is
\begin{align}
\mathcal L={}&\mathcal L_{\rm FM}
+\lambda_{\rm rec}\mathcal L_{\rm rec}
+\lambda_{\rm pix}\mathcal L_{\rm rec,pix}
+\lambda_{\rm rank}\mathcal L_{\rm rank}
+\lambda_{\rm VI}D_{\rm KL}(q_\phi\|p_\psi)\nonumber\\
&+\lambda_{\rm shrink}\mathcal L_{\rm shrink}
+\lambda_{\rm same}\mathcal L_{\rm same}.
\label{eq:full-implemented-objective}
\end{align}
The continuous variational term uses the fixed-scale divergence from
Eq.~\ref{eq:fixedkl}. Its numerical coefficient should therefore be
interpreted together with the explicit factor $1/(2\sigma_z^2)$ rather than as
an indication that the residual is negligible.

% \paragraph{Staged curriculum.}
% \textcolor{attentionorange}{
% After constructing the semantic memory, we pretrain the prototype prior
% and $\varepsilon$-adapter (Stage I) and freeze them before transport training.
% Stage II follows three phases: (i) establish the deterministic
% semantic-conditioning path without sampling the variational residual;
% (ii) introduce the target-informed posterior and gradually increase
% the KL weight; and (iii) jointly refine the residual model and
% conditional transport objective. Algorithm~\ref{alg:facet-training}
% details the variational phase, while the preceding deterministic
% phase and KL warm-up follow this curriculum.
% At inference, the reported point-estimate experiments use the
% sequence-prior mean rather than residual-latent sampling.
% }

\paragraph{Staged curriculum.}
Stage I constructs the semantic memory and context map, fits the prototype
prior with Eq.~\ref{eq:prototype-objective}, and then trains the $\varepsilon$-adapter while continuing to update the prior; both are frozen before
transport training. Stage II has three phases:
(i) 2000 steps with $z_S=m_S$, without $D_{\rm KL}$ or $\mathcal L_{\rm same}$;
(ii) 2000 steps in which $\delta z_S$ is sampled from the posterior $q_\phi$
and $\lambda_{\rm VI}$ is ramped linearly over the first 1000 steps; and
(iii) 3000 steps at constant $\lambda_{\rm VI}$, in which $\delta z_S$ is
instead drawn from the prior $p_\psi$ with a probability ramped linearly
from 0 to 1 over the first 1500 steps, matching its use at inference.
In phases (ii) and (iii), a quarter of posterior draws use another image of
the same protein. Algorithm~\ref{alg:facet-training} summarizes the training process.
At inference, the reported point-estimate experiments use the prior-mean
state of Eq.~\ref{eq:prior-mean-state} and the reported predictive-sampling experiments instead draw the protein-specific residual from its prior using the
reparameterization as shown in Eq.~\ref{eq:prior-reparameterized-sampling-state}.

\subsection{Prototype construction details}
\label{app:prototypes}

\paragraph{Feature extraction and residualization.}
{The same frozen VAE encoder $E$ produces the
spatial target latent $z_1=E_Y(Y)=E(Y)\in\mathbb R^{4\times64\times64}$ and
context encoding
$E_C(C)=[E(C_{\rm nucl});E(C_{\rm ER});E(C_{\rm MT})]
\in\mathbb R^{12\times64\times64}$. Pooling and flattening give
\begin{align*}
\Phi_Y(Y)&=\operatorname{flatten}(\operatorname{AvgPool}_{4}(E_Y(Y)))
\in\mathbb R^{1024},\\
\Phi_C(C)&=\operatorname{flatten}(\operatorname{AvgPool}_{4}(E_C(C)))
\in\mathbb R^{3072}.
\end{align*}
The target latent $z_1=E_Y(Y)$ summarize training images, whereas
$E_C(C)$ remains available at inference.} An affine ridge predictor $b$ is fit
using training proteins only, and the per-image residual is
\begin{equation}
U_i=\Phi_Y(Y_i)-b\!\left(\Phi_C(C_i)\right).
\label{eq:app-residual}
\end{equation}
The held-out context-to-target explanatory analysis gives an $R^2$ of $0.377$.
This analysis motivates constructing sequence-dependent memory in the residual
feature space rather than directly in the raw target-feature space. {The same fitted ridge map also supplies the
context branch:
$z_C=f_C(C)=T_C(b(\Phi_C(C)))\in\mathbb R^{128}$.
Here $T_C$ standardizes the ridge predictions, applies a 128-component PCA,
and rescales the components. The ridge map and $T_C$ are fitted on training
data and then frozen; $b$ is not refitted for this branch.}

\paragraph{Protein-level averaging and projection.}
For a training protein $p$ with image set $\mathcal I_p$, we compute
\begin{equation}
e_p=T\!\left(
\frac{1}{|\mathcal I_p|}\sum_{i\in\mathcal I_p}U_i
\right),
\label{eq:app-protein-representation}
\end{equation}
where $T$ is a fixed standardized principal-component projection. Protein-level
averaging gives each training protein equal weight regardless of its number of
images. The reported instantiation uses a 128-dimensional representation that
retains $71.2\%$ of the explained variance.
{The projection $T$ is fitted on protein-averaged
residuals, distinct from the context-prediction projection $T_C$. For each
training image, the target-informed posterior receives the projected residual
$U_Y=T(\Phi_Y(Y)-b(\Phi_C(C)))\in\mathbb R^{128}$; paired images of the same
protein are projected identically.}

\paragraph{Prototype vocabulary and assignments.}
The memory contains $K=9$ prototype vectors based on UniProt-localization annotation. The annotation rule assigns each memory-construction protein
to a coarse semantic state. If
$\mathcal P_k$ is the set assigned to state $k$, the prototype is
\begin{equation}
V_k=\frac{1}{|\mathcal P_k|}
\sum_{p\in\mathcal P_k}e_p,
\qquad
\mathcal M=\{V_k\}_{k=1}^{K}.
\label{eq:app-prototype-centroids}
\end{equation}
All feature transformations, assignments, and centroids are estimated from
training proteins and frozen before evaluation. At inference, the prior
$p_\eta(R\mid S,\mathcal M)$ predicts a distribution over the prototype-valued
states, and the deployed readout is
\begin{equation}
m_S=\sum_{r=1}^{K}p_\eta(R=V_r\mid S,\mathcal M)V_r.
\label{eq:app-soft-readout}
\end{equation}
Target images and annotation-derived states are not provided to this prior at
inference.

\paragraph{Prototype posterior and prior parameterization.}
Both distributions are parameterized by three-layer MLPs with
width $512$ and SiLU activations (denoted by $g_\omega$ and $g_\eta$ below), producing logits over the $K=9$
prototypes:
\begin{align}\label{eq:proto-posterior-prior}
q_\omega(R=V_r\mid z_1,C,S,\mathcal M)
&=\operatorname{softmax}
\left(g_\omega([\widetilde{\Phi}_Y(Y);
\widetilde{\Phi}_C(C);\widetilde h_S])\right)_r,
\nonumber\\
p_\eta(R=V_r\mid S,\mathcal M)
&=\operatorname{softmax}
\left(g_\eta(\widetilde h_S)\right)_r,
\end{align}

with $\Phi_Y(Y)=\operatorname{flatten}(\operatorname{AvgPool}_{4}(z_1))=\operatorname{flatten}(\operatorname{AvgPool}_{4}(E_Y(Y)))$. The tilde denotes standardization using training-set statistics;
$\widetilde h_S$ is the standardized mean-pooled sequence embedding,
excluding special tokens. 
The annotation label $r\in\{1,\ldots,K\}$ identifies the corresponding
prototype $V_r\in\mathcal M$ and supervises both distributions;
the prototype vectors are not directly input to either network.

\subsection{Architecture and implementation details}
\label{app:setup}

% \subsubsection{Overview of architecture and conditioning}
% \label{sec:method-overview}

\paragraph{Representations and initialization.}
The frozen VAE encoder $E$ provides the target latent
$z_1=E_Y(Y)=E(Y)$ and, for HPA, the spatial context
$E_C(C)=[E(C_{\rm nucl});E(C_{\rm ER});E(C_{\rm MT})]$.
A fitted, frozen context map yields $z_C=f_C(C)$, while the
frozen protein encoder produces $h_S=E_S(S)$.
Target-image encoding supplies training targets and semantic features
but is not required at inference. Generation starts from
$z_0\sim p_0=\N(0,I)$ and decodes the generated latent using
the frozen decoder $D$.

% \paragraph{Conditioning and transport.}
% Figure~\ref{fig:facet-model} and Algorithm~\ref{alg:facet-inference} illustrate the FACET inference pipeline.
% The sequence representation $h_S$ queries the fixed memory
% $\mathcal M$ to obtain the soft readout $m_S$, which combines
% with the residual to form $z_S=m_S+\delta z_S$.
% Thus, $z_S$ is a derived semantic state rather than a separate
% latent variable in the probabilistic model.
% The spatial context encoding $E_C(C)$ is concatenated with the projected
% transport state to condition the LoRA-adapted $\varepsilon$-predictor,
% which retains its direct sequence pathway through $h_S$ and
% conditions its $\varepsilon$-adapter on $m_S$.
% The hierarchical head applies context-first modulation using
% $z_C$, followed by sequence refinement using $z_S$, and predicts
% a gated correction to the base velocity $v_L$.
% The resulting field is integrated in latent space and decoded
% to generate the image. Training-only posterior and
% target-dependent reconstruction paths are omitted at inference.

\paragraph{Frozen and trainable components.}
The pretrained error-predictor weights, VAE encoder and decoder, and protein sequence
encoder remain frozen.
The fitted ridge map $b$,
projections $T$ and $T_C$, and semantic memory are also frozen during transport
training. The predictor receives the spatial context encoding $E_C(C)$
concatenated channel-wise with $y_\tau$; the hierarchical context FiLM instead
receives the compact $z_C=f_C(C)$.
The prototype prior and $\varepsilon$-adapter are
pretrained and subsequently frozen. Transport training updates the continuous
residual components, context-first hierarchical head, and LoRA parameters
in the $\varepsilon$-predictor.
The reported instantiation adds 9.83M deployed trainable
parameters (0.588\% of the 1.67B-parameter instantiation);
training-only posterior parameters are reported separately
in Table~\ref{tab:facet-parameters}.

\paragraph{Sequence semantic state.}
For the default point-estimate experiments, we use the prior mean
\begin{equation}
z_S=m_S+\mu_p(h_S,m_S).
\label{eq:prior-mean-state}
\end{equation}
Predictive-sampling experiments instead draw the protein-specific residual from its prior using the reparameterization
\begin{equation}
z_S=m_S+\mu_p(h_S,m_S)+\sigma_z\,\varepsilon,\qquad \varepsilon\sim\N(0,I).
\label{eq:prior-reparameterized-sampling-state}
\end{equation}

\paragraph{Fixed settings.}
The following values are fixed reproducibility settings for the reported FACET
instantiation and \textit{are not claimed to be individually optimal}. For the reported experiments, we fix the residual-scale constants to
$c_\sigma=0.05$ and $c_\mu=0.25$.
\begin{table}[!h]
\caption{Selected implementation settings for the FACET instantiation used
in the reported experiments; component-level ablations are reported separately.}
\label{tab:facet-configuration}
\centering
\setlength{\tabcolsep}{5pt}
\begin{tabular}{@{}ll@{}}
\toprule
Setting & Value \\
\midrule
Semantic memory vocabulary
    & $K=9$ fixed UniProt localization categories \\
Residual representation
    & 128 dimensions \\
% Residual scaling
%     & $c_\sigma=0.05,\; c_\mu=0.25$ \\
Interpolant strength
    & $\kappa=0.5$ \\
Correction head
    & 2 residual blocks \\
LoRA adaptation
    & rank $16$ \\
Inference budget
    & 25 NFE \\
\bottomrule
\end{tabular}

\end{table}

\paragraph{Relation to residual and additive parameterizations.}
FACET employs residual operations at four levels:
context-residualized target features, a continuous semantic residual around
the coarse memory, a residual correction to the
pretrained noise estimate, and a correction to the base velocity field.
These operations serve distinct purposes within the factorized probabilistic
model. In particular, the deployed semantic state $z_S=m_S+\delta z_S$ is
additive, but the conditional generator is not restricted to an additive
decomposition of independent context and sequence functions. For example,
sequential FiLM modulation introduces multiplicative interactions through
{the product
$(1+\gamma_S(z_S))\odot(1+\gamma_C(z_C))$}. Moreover,
{the base velocity $v_L$ is itself conditioned
jointly on $S$ and $C$, and} the composed velocity
$v_\Theta=v_L+g(\tau)\Delta v$ is integrated along an evolving stochastic
trajectory, so additive velocity composition does not imply additive
decomposition of the generated image.

\subsection{Transport discretization and inference}
\label{app:flow}

\paragraph{Finite-schedule lookup.}
The continuous interpolant is mapped to the native variance-preserving schedule
using the exact normalization in Proposition~\ref{prop:vp}. For any pretrained
predictor that is defined at finite native noise levels, the projected state is
queried at
\begin{equation}
k(\tau)=\arg\min_k|\tau_k-\tau|.
\label{eq:app-schedule-lookup}
\end{equation}
The state projection itself preserves the variance-preserving identity exactly;
the finite-index lookup is the remaining schedule-specific approximation.
{Here $\tau_k$ denotes the continuous coordinate
associated with native VP noise index $k$; lookup selects a represented native
level rather than a continuously adjustable predictor timestep.}

\paragraph{Inference-time transport.}
At inference, the model computes $m_S$ from the protein sequence, obtains $z_S$ from the sequence-prior, and evaluates the
composed velocity
\begin{equation}
v_\Theta=v_L+g(\tau)\Delta v.
\label{eq:app-composed-velocity}
\end{equation}
The latent trajectory is integrated with the reported Euler-Maruyama sampler~\citep{albergo2023stochastic,song2021sde}.
{The reported uniform sampling range is
$\tau_{\min}=0.05$ to $\tau_{\max}=1$, with
$\Delta\tau=(\tau_{\max}-\tau_{\min})/n$.
The primary operating point uses $n=25$ network function evaluations and
$n-1$ state updates; the final evaluation occurs at
$\tau_{\max}-\Delta\tau$ and emits the base clean-latent estimate
$\hat z_1^L$, which is decoded as $\hat Y=D(\hat z_1^L)$.
Initialization at nonzero $\tau_{\min}$ uses a noise-derived latent and is
an approximation to the target-dependent interpolant marginal.}
The deterministic ODE variant, integrated using Heun's method~\citep{karras2022elucidating}, is evaluated only as a sampling ablation (Tab.~\ref{tab:opencell-samplers}).

\section{Algorithms}
Algorithms~\ref{alg:facet-training} and~\ref{alg:facet-inference} summarize FACET training and inference, respectively. Here, $\operatorname{AdaptedPredictor}(\cdot)$ denotes the pretrained $\varepsilon$-predictor augmented with LoRA and the $\varepsilon$-adapter, reusing the sequence embedding $h_S$ and memory readout $m_S$ computed during conditioning. Fig.~\ref{fig:facet-model} depicts the inference pipeline.
\input{architecture}

% \subsection{Training algorithm}
\begin{algorithm}[p]
\caption{\FACET{} training}
\label{alg:facet-training}
\small
\begin{algorithmic}[1]
\Require Training data $\mathcal D=\{(S_i,C_i,Y_i,r_i)\}$, frozen pretrained
$\varepsilon$-predictor, encoders and decoder, loss weights and curriculum
\Ensure Trained \FACET{} model and fixed memory $\mathcal M$
%%%%%%%%%%%%%

\Statex \textbf{Stage I: Memory, context map, and prior pretraining}

\State {
    Extract $\{\Phi_Y(Y_i),\Phi_C(C_i)\}_i$
     following Appendix~\ref{app:prototypes}
}
\State {
    Fit ridge map $b$; compute $U_i$ using Eq.~\ref{eq:app-residual}
}
\State {
    Fit $T$; construct and freeze $\mathcal M$
    using Eqs.~\ref{eq:app-protein-representation}--\ref{eq:app-prototype-centroids}
}
\State {
    Fit $T_C$; freeze $f_C=T_C\circ b\circ\Phi_C$ 
    following Appendix~\ref{app:prototypes}
}
\For{minibatches $(S_i,C_i,Y_i,r_i)$}
    \State {
        $z_{1,i}\gets E_Y(Y_i)$;
        evaluate $q_\omega(\cdot\mid z_{1,i},S_i,C_i;\mathcal M)$
        and $p_\eta(\cdot\mid S_i;\mathcal M)$ using Eq.~\ref{eq:proto-posterior-prior}
        % $q_\omega(R\mid\Phi_Y(Y),\Phi_C(C),E_S(S))$ % q_\omega(R\mid z_1,S,C,\mathcal M) \,\middle\|\, p_\eta(R\mid S,\mathcal M).
    }
    \State {
        Update $(\omega,\eta)$ using
        $\mathcal L_{\rm proto}$ in Eq.~\ref{eq:prototype-objective}
    }
\EndFor
\State {
    Pretrain the $\varepsilon$-adapter while updating $(\omega,\eta)$
    following Appendix~\ref{app:training}
}
\State {
    Freeze the $\varepsilon$-adapter and $p_\eta$
}

\vspace{7pt}
\Statex \textbf{Stage II: Residual and transport training}
\Statex \textit{Trainable: LoRA, correction head, and
variational residual $(\phi,\psi)$.
Curriculum: Appendix~\ref{app:training}.}
% \Statex Trainable: LoRA on $\varepsilon$-predictor, hierarchical correction head,
%     and variational residual $(\phi,\psi)$; Stage I components frozen.
% \Statex {Phase schedule and $\lambda_{\rm VI}$ warm-up
%     follow Appendix~\ref{app:training}.}

\For{transport-training minibatches}
    \State Sample $(S,C,Y)$ and a same-protein image $(C^{(j)},Y^{(j)})$
    \State $z_1\gets E_Y(Y)$; $z_C\gets f_C(C)$; $h_S\gets E_S(S)$
    \State $m_S\gets\sum_{r=1}^{K}p_\eta(R=V_r\mid S,\mathcal M)V_r$
        \Comment{Soft readout}
    
    \If{{phase (i)}}
        \State {$\delta z_S\gets 0$}
    \Else
        \State $U_Y\gets T(\Phi_Y(Y)-b(\Phi_C(C)))$;
            $U_Y^{(j)}\gets T(\Phi_Y(Y^{(j)})-b(\Phi_C(C^{(j)})))$
        \State $\bar{\mu}_q\gets\mu_q(U_Y,m_S)$;
               $\bar{\mu}_q^{(j)}\gets\mu_q(U_Y^{(j)},m_S)$;
               $\bar{\mu}_p\gets\mu_p(h_S,m_S)$
        \If{{phase (iii) and $\mathrm{Bernoulli}(\rho)$}}
            \Comment{{$\rho$ ramps from 0 to 1}}
            \State {$\mu\gets\bar{\mu}_p$}
        \ElsIf{{$\mathrm{Bernoulli}(0.25)$}}
            \State {$\mu\gets\bar{\mu}_q^{(j)}$}
                \Comment{{Same-protein swap}}
        \Else
            \State {$\mu\gets\bar{\mu}_q$}
        \EndIf
        \State {$\delta z_S\gets\mu+\sigma_z\varepsilon$,
            $\varepsilon\sim\mathcal N(0,I)$}
    \EndIf
    \State $z_S\gets m_S+\delta z_S$

    % \Statex \textbf{Stochastic interpolant}
    % \Statex \vspace{1pt}
    \vspace{6pt}
    \Statex \hspace{\algorithmicindent}\textbf{\textit{Stochastic interpolant}}
    \State Sample native index $k$ with $\tau_k\ge\tau_{\min}$
        and $Q_{\tau_k}>0$; set $\tau\gets\tau_k$
    \State Sample $z_0,\xi\overset{\rm i.i.d.}{\sim}\mathcal N(0,I)$
    \State $x_\tau\gets A_\tau z_1+B_\tau z_0+G_\tau\xi$
    \State $\epsilon_\tau\gets(B_\tau z_0+G_\tau\xi)/Q_\tau$;
           $u_\tau^\star\gets z_1+\dot Q_\tau\epsilon_\tau$

    % \Statex 
    \vspace{6pt}
    \Statex \hspace{\algorithmicindent}\textbf{\textit{Conditional transport}}
    \State $y_\tau\gets x_\tau/N_\tau$
    \State {$(\hat\epsilon_L,H_k)\gets
        \operatorname{AdaptedPredictor}(y_\tau,k,S,E_C(C),m_S)$}
        \Comment{Adapted $\varepsilon$-predictor}
        % \Comment{LoRA-adapted $\varepsilon$-predictor with $\varepsilon$-adapter}
    \State $\hat z_1^L\gets
        (x_\tau-Q_\tau\hat\epsilon_L)/\max(A_\tau,\epsilon_{\rm num})$
    \State $v_L\gets\hat z_1^L+\dot Q_\tau\hat\epsilon_L$
    \State $v_\Theta\gets v_L+g(\tau)\Delta v(H_k,z_C,z_S,\tau)$
    \State $\hat z_1^\theta\gets\hat z_1^L+g(\tau)\Delta v(H_k,z_C,z_S,\tau)$
        \Comment{Training only}

    % \Statex \vspace{1pt}
    \vspace{6pt}
    \Statex \hspace{\algorithmicindent}\textbf{\textit{Losses and optimization}}
    \State Compute $\mathcal L_{\rm FM}$, $\mathcal L_{\rm rec}$,
        $\mathcal L_{\rm rec,pix}$, and $\mathcal L_{\rm shrink}$
        using Eqs.~\ref{eq:flow-objective}--\ref{eq:regularization-objectives}
        {($\mathcal L_{\rm rec,pix}$ on every fourth step)}
    \State {For one example per minibatch, sample $n$
        negatives (Appendix~\ref{app:training})}
    \State {Evaluate each at the same $(x_\tau,\tau,C)$ with
        its own $z_S$ (Eq.~\ref{eq:prior-mean-state}; $z_S=m_S$ in phase (i))}
    \State Compute $\mathcal L_{\rm rank}$ using Eq.~\ref{eq:ranking-objective}
    \If{{phase (ii) or (iii)}}
        \State Compute $D_{\rm KL}(q_\phi\|p_\psi)$ from $(\bar\mu_q,\bar\mu_p)$
            using Eq.~\ref{eq:fixedkl}
        \State Compute $\mathcal L_{\rm same}$ from
            {$(\bar\mu_q,\bar\mu_q^{(j)})$}
            using Eq.~\ref{eq:regularization-objectives}
    \EndIf
    \State Form $\mathcal L$ using Eq.~\ref{eq:full-implemented-objective}
    \State Update {LoRA, the correction head, and,
        from phase (ii), $(\phi,\psi)$} using $\nabla\mathcal L$

\EndFor

\State \Return Trained model and $\mathcal M$

%%%%%%%%%%%%%
\end{algorithmic}
\end{algorithm}

\begin{algorithm}[!h]
\caption{\FACET{} stochastic inference}
\label{alg:facet-inference}
\begin{algorithmic}[1]
\Require Protein sequence $S$, cellular context $C$,
fixed memory $\mathcal M=\{V_r\}_{r=1}^{K}$,
trained model, sampling parameters
$(n,\tau_{\min},\tau_{\max},\eta_0)$
\Ensure Generated fluorescence image $\hat Y$

\Statex \textbf{Conditioning}
\State $z_C\gets f_C(C)$; $h_S\gets E_S(S)$
\State $m_S\gets\sum_{r=1}^{K}p_\eta(R=V_r\mid S,\mathcal M)V_r$
    \Comment{Soft memory readout}
\State $\delta z_S\gets\mu_p(h_S,m_S)$ \Comment{Default prior mean}
\State Optional predictive sampling: $\delta z_S\gets\mu_p(h_S,m_S)+\sigma_z\varepsilon$,
    $\varepsilon\sim\mathcal N(0,I)$
\State $z_S\gets m_S+\delta z_S$

\Statex \textbf{Initialization}
\State Sample $z_0,\xi\overset{\mathrm{i.i.d.}}{\sim}\mathcal N(0,I)$
\State $x_{\tau_{\min}}$$\gets B_{\tau_{\min}}z_0+G_{\tau_{\min}}\xi$
    \Comment{{$A_{\tau_{\min}}z_1$ omitted}}
\State $\Delta\tau\gets(\tau_{\max}-\tau_{\min})/n$

\Statex \textbf{Stochastic transport}
\For{$j=0,\ldots,n-1$}
    \State $\tau\gets\tau_{\min}+j\Delta\tau$
    \State $y_\tau\gets$ {$x_\tau$}$/N_\tau$
        \Comment{VP projection}
    \State $k\gets\arg\min_{\ell}|\tau_\ell-\tau|$
        \Comment{{Eq.~\ref{eq:schedule-lookup}}}
    \State $(\hat\epsilon_L,H_k)\gets
        \operatorname{AdaptedPredictor}(y_\tau,k,S,
        E_C(C),m_S)$
        \Comment{Adapted $\varepsilon$-predictor}
    \State $\hat z_1^L\gets
        (x_\tau-Q_\tau\hat\epsilon_L)/\max(A_\tau,\epsilon_{\rm num})$
    \State $v_L\gets\hat z_1^L+\dot Q_\tau\hat\epsilon_L$
    \State $v_\Theta\gets v_L+g(\tau)\Delta v(H_k,z_C,z_S,\tau)$
        \Comment{Apply hierarchical correction}
    \If{$j<n-1$}
        \State $s_\Theta\gets-\hat\epsilon_L/Q_\tau$
        \State $\eta_\tau\gets\eta_0\tau(1-\tau)$
        \State Sample $\zeta_j\sim\mathcal N(0,I)$
        \State $x_{\tau+\Delta\tau}\gets x_\tau+(v_\Theta+\eta_\tau s_\Theta)\Delta\tau+\sqrt{2\eta_\tau\Delta\tau}\,\zeta_j$
            \Comment{Euler-Maruyama}
    \EndIf
\EndFor

\Statex \textbf{Emission}
\State $\hat Y\gets D(\hat z_1^L)$
    \Comment{Last evaluation, $\tau=\tau_{\max}-\Delta\tau$}
\State \Return $\hat Y$
\end{algorithmic}
\end{algorithm}

% \clearpage % Keep the algorithm floats ahead of the proofs.

\section{Proofs, Derivations, and other theoretical properties}
\label{app:proofs}

\subsection{Proof of Proposition~\ref{prop:elbo}}
\label{app:proof-elbo}

Since $R$ is a prototype-valued discrete variable supported on $\mathcal M$,
\begin{equation}
p_\Theta(z_1\mid S,C,\mathcal M)
=\sum_{R\in\mathcal M}\int
p_\Theta(z_1,R,\delta z_S\mid S,C,\mathcal M)\,
\dd\delta z_S.
\end{equation}
Multiplying and dividing the integrand by
$q(R,\delta z_S\mid z_1,S,C,\mathcal M)$ and applying Jensen's
inequality gives
\begin{align}
\log p_\Theta(z_1\mid S,C,\mathcal M)
\geq
\mathbb E_q\left[
\log
\frac{
p_\Theta(z_1,R,\delta z_S\mid S,C,\mathcal M)}
{q(R,\delta z_S\mid z_1,S,C,\mathcal M)}
\right].
\end{align}
Substituting the generative factorization in Eq.~\ref{eq:factor} and
the posterior factorization from Proposition~\ref{prop:elbo}, the
log-ratio decomposes as

\begin{align}
&\log p_\theta(z_1\mid S,C,R,\delta z_S)
+\log\frac{p_\eta(R\mid S,\mathcal M)}
{q_\omega(R\mid z_1,S,C,\mathcal M)}
\nonumber\\
&\qquad
+\log\frac{p_\psi(\delta z_S\mid S,R)}
{q_\phi(\delta z_S\mid z_1,S,C,R,\mathcal M)}.
\end{align}
Taking expectations therefore yields
\begin{align*}
\log p_\Theta(z_1\mid S,C,\mathcal M)
\geq{}&
\mathbb E_q\!\left[
\log p_\theta(z_1\mid S,C,R,\delta z_S)
\right]\\
&-D_{\mathrm{KL}}(q_\omega\|p_\eta)
-\mathbb E_{q_\omega}
D_{\mathrm{KL}}(q_\phi\|p_\psi),
\end{align*}
which is Eq.~\ref{eq:elbo}.

\subsection{Derivation of the base latent estimate and velocity}
\label{app:base-velocity-derivation}

We derive the base clean-latent estimate and velocity in
Eq.~\ref{eq:base-velocity} from the Gaussian-reparameterized
stochastic interpolant. Recall that
\begin{equation}
x_\tau
=
A_\tau z_1+B_\tau z_0+G_\tau\xi,
\qquad
A_\tau=\tau,
\label{eq:app-original-interpolant}
\end{equation}
where $z_0,\xi\overset{\mathrm{i.i.d.}}{\sim}\mathcal N(0,I)$
are independent of the target latent $z_1$.
Define
\begin{equation}
Q_\tau=\sqrt{B_\tau^2+G_\tau^2},
\qquad
\epsilon_\tau
=
\frac{B_\tau z_0+G_\tau\xi}{Q_\tau}.
\label{eq:app-effective-noise}
\end{equation}
For $Q_\tau>0$, independence and Gaussianity imply
\begin{equation}
\epsilon_\tau\sim\mathcal N(0,I),
\qquad
x_\tau=A_\tau z_1+Q_\tau\epsilon_\tau.
\label{eq:app-reparameterized-interpolant}
\end{equation}
% \textcolor{attentionorange}{The endpoint at which $Q_\tau=0$ is understood by continuity.}

\paragraph{Clean-latent estimation.}
For $A_\tau>0$, rearranging
Eq.~\ref{eq:app-reparameterized-interpolant} gives
\begin{equation}
z_1
=
\frac{x_\tau-Q_\tau\epsilon_\tau}{A_\tau}.
\label{eq:app-exact-clean-latent}
\end{equation}
The pretrained $\varepsilon$-predictor is evaluated on the
variance-preserving projection of $x_\tau$ at the selected
native noise index, producing a noise estimate
$\hat\epsilon_L$. Substituting this estimate for the
unknown effective noise yields
\begin{equation}
\hat z_1^L
=
\frac{x_\tau-Q_\tau\hat\epsilon_L}{A_\tau}.
\label{eq:app-clean-estimate-unclipped}
\end{equation}
To avoid division by a vanishing coefficient near
$\tau=0$, the implemented estimator uses
\begin{equation}
\boxed{
\hat z_1^L
=
\frac{x_\tau-Q_\tau\hat\epsilon_L}
{\max(A_\tau,\epsilon_{\mathrm{num}})}.
}
\label{eq:app-clean-estimate}
\end{equation}
The unclipped expression follows exactly from the
reparameterized interpolant when the effective noise
is known; the clipped expression is its numerically
stabilized estimator.

\paragraph{Base velocity.}
The original interpolant has pathwise derivative
\begin{equation}
\dot x_\tau
=
\dot A_\tau z_1
+\dot B_\tau z_0
+\dot G_\tau\xi.
\label{eq:app-original-path-derivative}
\end{equation}
The effective noise $\epsilon_\tau$ generally depends
on $\tau$ when constructed from fixed draws of
$z_0$ and $\xi$. Thus, differentiating the
reparameterized expression while holding
$\epsilon_\tau$ fixed is not, in general, the
pathwise derivative of the original sampled
trajectory. Instead, the relevant conditional
velocity is obtained by averaging
Eq.~\ref{eq:app-original-path-derivative}
over the unresolved Gaussian components.

Because $(z_0,\epsilon_\tau)$ and
$(\xi,\epsilon_\tau)$ are jointly Gaussian and
zero-mean, with
$\operatorname{Cov}(\epsilon_\tau,\epsilon_\tau)=I$, their
conditional expectations are
\begin{align}
\mathbb E[z_0\mid\epsilon_\tau]
&=
\operatorname{Cov}(z_0,\epsilon_\tau)
\operatorname{Cov}(\epsilon_\tau,\epsilon_\tau)^{-1}
\epsilon_\tau
=
\frac{B_\tau}{Q_\tau}\epsilon_\tau,
\\
\mathbb E[\xi\mid\epsilon_\tau]
&=
\operatorname{Cov}(\xi,\epsilon_\tau)
\operatorname{Cov}(\epsilon_\tau,\epsilon_\tau)^{-1}
\epsilon_\tau
=
\frac{G_\tau}{Q_\tau}\epsilon_\tau.
\end{align}
Since $z_1$ is independent of the Gaussian noise
variables, conditioning additionally on $z_1$
does not change these expectations. Consequently,
\begin{align}
\mathbb E[\dot x_\tau\mid z_1,\epsilon_\tau]
&=
\dot A_\tau z_1
+\dot B_\tau
  \mathbb E[z_0\mid\epsilon_\tau]
+\dot G_\tau
  \mathbb E[\xi\mid\epsilon_\tau]
\nonumber\\
&=
\dot A_\tau z_1
+
\frac{B_\tau\dot B_\tau+
      G_\tau\dot G_\tau}{Q_\tau}
\epsilon_\tau.
\label{eq:app-conditional-velocity-intermediate}
\end{align}
Differentiating
$Q_\tau=\sqrt{B_\tau^2+G_\tau^2}$ gives
\begin{equation}
\dot Q_\tau
=
\frac{B_\tau\dot B_\tau+
      G_\tau\dot G_\tau}{Q_\tau}.
\end{equation}
Therefore,
\begin{equation}
\mathbb E[\dot x_\tau\mid z_1,\epsilon_\tau]
=
\dot A_\tau z_1+\dot Q_\tau\epsilon_\tau.
\label{eq:app-conditional-velocity}
\end{equation}
For the chosen interpolation coefficient
$A_\tau=\tau$, we have $\dot A_\tau=1$,
so the conditional velocity becomes
\begin{equation}
u_\tau^\star
=
z_1+\dot Q_\tau\epsilon_\tau.
\label{eq:app-conditional-flow-target}
\end{equation}
This is the flow-matching target used in
Appendix~\ref{app:training}.

Replacing the unknown clean latent and effective
noise with their predictor-based estimates gives
the base velocity
\begin{equation}
\boxed{
v_L
=
\hat z_1^L+\dot Q_\tau\hat\epsilon_L.
}
\label{eq:app-derived-base-velocity}
\end{equation}
Together,
Eqs.~\ref{eq:app-clean-estimate} and
\ref{eq:app-derived-base-velocity} recover
the two expressions in Eq.~\ref{eq:base-velocity}.

\paragraph{Scope of the derivation.}
The Gaussian reparameterization and conditional-velocity identities above
are exact under the stated noise assumptions. The resulting predictor-based
field is an estimate: the variance-preserving projection establishes
compatibility with the pretrained predictor's noise family, while
nearest-index lookup approximates its finite schedule. In particular,
$\hat{\varepsilon}_L$ estimates the conditional noise rather than the realized
$\varepsilon_\tau$. Flow matching therefore trains the hierarchical correction
against the realized target $u_\tau^\star$ to capture the conditionally
predictable residual between the base field and target velocity.

\subsection{Derivation and interpretation of the auxiliary latent estimate}
\label{app:auxiliary-clean-derivation}

The Gaussian-reparameterized interpolant is
\begin{equation}
x_\tau=A_\tau z_1+Q_\tau\epsilon_\tau,
\qquad A_\tau=\tau.
\end{equation}
As derived above, conditioning the interpolant's pathwise
derivative on $(z_1,\epsilon_\tau)$ yields the conditional
velocity target
\begin{equation}
u_\tau^\star
=\dot A_\tau z_1+\dot Q_\tau\epsilon_\tau
=z_1+\dot Q_\tau\epsilon_\tau,
\end{equation}
where $\dot A_\tau=1$. Accordingly, the base velocity and
hierarchically corrected velocity are
\begin{equation}
v_L=\hat z_1^L+\dot Q_\tau\hat\epsilon_L,
\qquad
v_\Theta=v_L+g(\tau)\Delta v.
\label{eq:app-velocity-decomposition}
\end{equation}

\paragraph{Auxiliary reconstruction objective.}
Flow matching supervises the corrected velocity, while an auxiliary
reconstruction objective provides direct supervision of clean-latent
accuracy through the same correction head. To construct the
\textit{training-time} auxiliary latent estimate, we assign the velocity
correction to the clean-latent component while retaining the
adapted noise estimate $\hat\epsilon_L$. This fixed-noise choice
uses the existing $\varepsilon$-predictor without introducing an additional
one. Inverting the velocity decomposition then gives
\begin{align}
\hat z_1^\theta
&:=\frac{v_\Theta-\dot Q_\tau\hat\epsilon_L}
{\dot A_\tau}
\nonumber\\
&=v_L+g(\tau)\Delta v-\dot Q_\tau\hat\epsilon_L
\nonumber\\
&=\hat z_1^L+g(\tau)\Delta v,
\label{eq:app-auxiliary-clean}
\end{align}
where the second line uses $\dot A_\tau=1$.

% The fixed-noise decomposition is a modeling choice, not a unique
% inversion of the corrected velocity: it does not additionally
% enforce
% $x_\tau=A_\tau\hat z_1^\theta+Q_\tau\hat\epsilon_L$.
% The resulting auxiliary loss in the latent-space is
% \begin{equation}
% \mathcal L_{\mathrm{rec}}
% =
% \mathbb E\left[
% \left\|
% \underbrace{\hat z_1^L+g(\tau)\Delta v}_{\hat z_1^\theta}
% -z_1
% \right\|_2^2
% \right],
% \label{eq:app-auxiliary-rec-loss}
% \end{equation}
% and in the pixel-space is
% \begin{equation}
% \mathcal L_{\mathrm{rec,pix}}
% =
% \mathbb E\left[
% \left\|
% D(\underbrace{\hat z_1^L+g(\tau)\Delta v}_{\hat z_1^\theta})
% -D(z_1)
% \right\|_2^2
% \right],
% \label{eq:app-auxiliary-rec-loss}
% \end{equation}
% with frozen decoder $D$. These complement flow matching by directly supervising
% clean-latent accuracy.
% The auxiliary latent estimate $\hat z_1^\theta$ is used only
% during training; inference integrates $v_\Theta$.

The fixed-noise decomposition is a modeling choice, not a unique
inversion of the corrected velocity: it does not additionally enforce
$x_\tau=A_\tau\hat z_1^\theta+Q_\tau\hat\epsilon_L$.
The resulting auxiliary objectives in latent and pixel space are
\begin{equation}
\mathcal L_{\mathrm{rec}}
=
\mathbb E\left[
\left\|\hat z_1^\theta-z_1\right\|_2^2
\right],
\label{eq:app-auxiliary-rec-loss}
\end{equation}
and
\begin{equation}
\mathcal L_{\mathrm{rec,pix}}
=
\mathbb E\left[
\left\|D(\hat z_1^\theta)-D(z_1)\right\|_2^2
\right],
\label{eq:app-auxiliary-pixel-rec-loss}
\end{equation}
respectively, where $D$ is the frozen decoder.
During training, these auxiliary objectives complement flow matching by
providing direct latent- and pixel-space supervision.

%%%%%
We use $D(z_1)$ rather than $Y$ as the pixel-space target so that the
auxiliary objective remains anchored to the same clean latent endpoint
$z_1$, without requiring the learned correction to compensate for
reconstruction error of the frozen decoder. Moreover, letting
$r=\hat z_1^\theta-z_1$, a first-order expansion gives
\begin{equation}
D(\hat z_1^\theta)-D(z_1)
\approx J_D(z_1)r,
\qquad
\mathcal L_{\mathrm{rec,pix}}
\approx
\mathbb E\!\left[
% \frac{1}{P}
r^\top J_D(z_1)^\top J_D(z_1)r
\right].
\end{equation}
% where $P=H\times W\times C$ is the total number of decoded image elements, for image height $H$, width $W$, and number of channels $C$. 
Thus, unlike latent-space MSE, which penalizes latent directions uniformly, the pixel-space objective locally weights latent discrepancies according to their sensitivity under the frozen decoder.
%%%%%

The auxiliary estimate $\hat z_1^\theta$ is used only to construct these
objectives; inference directly integrates $v_\Theta$.

\subsection{Derivation of the stochastic sampler}
\label{app:stochastic-sampler-derivation}

\paragraph{Score estimation.}
As derived above, the interpolant admits the Gaussian
reparameterization
$x_\tau=A_\tau z_1+Q_\tau\epsilon_\tau$,
where $\epsilon_\tau\sim\mathcal N(0,I)$.
The Gaussian-denoising score identity
\citep{albergo2023stochastic} gives
\begin{equation}
\nabla_x\log p_\tau(x\mid S,C)
=
-\frac{\mathbb E[\epsilon_\tau\mid x_\tau=x,S,C]}
{Q_\tau}.
\end{equation}
Motivated by this identity, we construct an approximate score from the adapted
noise prediction:
\begin{equation}
s_\Theta=-\frac{\hat\epsilon_L}{Q_\tau}.
\label{eq:app-score-estimate}
\end{equation}

\paragraph{Stochastic dynamics.}
The stochastic-interpolant framework
\citep{albergo2023stochastic} shows that an exact
probability-flow velocity and its corresponding score can
define an SDE with the same time-dependent marginals,
provided the initialization and regularity conditions hold.
For \FACET{}, we use the learned corrected velocity
$v_\Theta$ and approximate score $s_\Theta$ in the
following SDE:
\begin{equation}
\mathrm dx_\tau
=
\left(v_\Theta+\eta_\tau s_\Theta\right)\mathrm d\tau
+\sqrt{2\eta_\tau}\,\mathrm dW_\tau,
\label{eq:app-facet-sde}
\end{equation}
where $\eta_\tau=\eta_0\tau(1-\tau)$ and $W_\tau$
denotes standard Brownian motion.

\paragraph{Euler-Maruyama sampling.}
For $\tau_j=\tau_{\min}+j\Delta\tau$, the Brownian increment
satisfies
$\Delta W_j\overset{d}{=}\sqrt{\Delta\tau}\,\zeta_j$,
where $\zeta_j\sim\mathcal N(0,I)$.
Using Euler-Maruyama sampling
\citep{albergo2023stochastic,higham2001algorithmic}, we obtain the update
\begin{equation}
\boxed{
x_{j+1}
=
x_j+
\left(v_\Theta+\eta_{\tau_j}s_\Theta\right)\Delta\tau
+\sqrt{2\eta_{\tau_j}\Delta\tau}\,\zeta_j.
}
\label{eq:app-euler-maruyama}
\end{equation}
All fields are evaluated at $(x_j,\tau_j)$, and each update
uses the corrected velocity $v_\Theta$. Initialization and
final emission follow Algorithm~\ref{alg:facet-inference}.

\subsection{Fixed-scale Kullback--Leibler divergence}
\label{app:proof-fixed-kl}

For two Gaussian distributions with shared isotropic covariance,
\begin{equation}
q_\phi=\N(\mu_q,\sigma_z^2I),
\qquad
p_\psi=\N(\mu_p,\sigma_z^2I),
\end{equation}
let $d_z$ denote the residual dimension. The KL divergence between two
Gaussians is
\begin{equation}
\begin{aligned}
D_{\mathrm{KL}}(q_\phi\|p_\psi)
=\frac{1}{2}\Big[
&\operatorname{tr}\!\left(\Sigma_p^{-1}\Sigma_q\right)
+(\mu_p-\mu_q)^\top\Sigma_p^{-1}(\mu_p-\mu_q)\\
&-d_z+\log\frac{\det\Sigma_p}{\det\Sigma_q}
\Big].
\end{aligned}
\end{equation}
Here $\Sigma_q=\Sigma_p=\sigma_z^2I$, so the log-determinant ratio
vanishes and
$\operatorname{tr}(\Sigma_p^{-1}\Sigma_q)=d_z$, which cancels the
$-d_z$ term. Since $\Sigma_p^{-1}=\sigma_z^{-2}I$, the remaining
quadratic term gives
\begin{equation}
D_{\mathrm{KL}}(q_\phi\|p_\psi)
=
\frac{1}{2\sigma_z^2}
(\mu_p-\mu_q)^\top(\mu_p-\mu_q)
=
\frac{\|\mu_q-\mu_p\|_2^2}{2\sigma_z^2},
\end{equation}
as stated in Eq.~\ref{eq:fixedkl}.

\subsection{Proof of Proposition~\ref{prop:vp}}
\label{app:proof-vp}

Because $z_0$ and $\xi$ are independent standard Gaussian variables,
\begin{equation}
B_\tau z_0+G_\tau\xi
\sim\N\!\left(0,(B_\tau^2+G_\tau^2)I\right).
\end{equation}
Thus, with $Q_\tau=\sqrt{B_\tau^2+G_\tau^2}$,
\begin{equation}
\epsilon_\tau
=\frac{B_\tau z_0+G_\tau\xi}{Q_\tau}
\sim\N(0,I),
\qquad
x_\tau=A_\tau z_1+Q_\tau\epsilon_\tau.
\end{equation}
Dividing by $N_\tau=\sqrt{A_\tau^2+Q_\tau^2}$ gives
\begin{equation}
y_\tau
=\frac{x_\tau}{N_\tau}
=a_\tau z_1+\sigma_\tau\epsilon_\tau,
\qquad
a_\tau=\frac{A_\tau}{N_\tau},
\quad
\sigma_\tau=\frac{Q_\tau}{N_\tau}.
\end{equation}
Therefore,
\begin{equation}
a_\tau^2+\sigma_\tau^2
=\frac{A_\tau^2+Q_\tau^2}{N_\tau^2}=1,
\end{equation}
which proves the proposition.
% ============================================================
% Appendix: Additional Experimental Details and Results
% ============================================================

% \section{Additional Experimental Details and Results}
% \label{app:additional-results}

% This appendix provides detailed experimental setup information, metric
% definitions, evaluation protocols, supporting analyses, and additional
% qualitative examples. Unless otherwise stated, all experiments use the same
% checkpoints, data partitions, conditioning inputs, and sampling procedures as
% the main paper.

% ============================================================
\section{Experimental Setup and Data Preparation}
\label{app:setup}
% This appendix section provides detailed experimental setup information. 
% ============================================================

\subsection{Human Protein Atlas benchmark}
\label{app:hpa}

The HPA benchmark is derived from the Human Protein Atlas
\citep{thul2017hpa}. Each example contains a protein amino-acid sequence, a
target protein-fluorescence image, and three cellular-context channels
corresponding to the nucleus, endoplasmic reticulum (ER), and microtubules
(MT). Images are cropped to $1024\times1024$ pixels and resized to
$256\times256$ pixels before the frozen variational autoencoder (VAE) maps
them to $4\times64\times64$ latent representations.

{The released HPA training set contains 87,763	stored images from 12,732 proteins 
% (of which 2,467 images from 349 proteins are used for validation), 
and the test split contains 720 stored images from 100 proteins}\footnote{\url{https://virtualcellmodels.cziscience.com/dataset/hpa-subcellular-section}}.
Following \cite{zheng2025celldiff}, we use a maximum tokenized sequence length
of 2048. This filtering removes 24 images from three proteins, yielding the
696-image, 97-protein evaluation set used throughout the paper. The longest
retained sequence has 1938 residues, whereas the shortest excluded sequence
has 2157 residues; therefore, the filtering boundary does not remove sequences close to the cutoff. 
%
% For FACET module adaptation and memory building, we cache a random subset of the
% HPA training split, drawn by adding whole proteins in a fixed random order until
% the image budget is reached and keeping every image of each selected protein.
% This yields 24{,}001 images from 3{,}491 of the 12{,}732 HPA training proteins;
% the UniProt annotation filter is applied later, when prototypes are assigned.
% Of these, 3{,}007 proteins carry a localization annotation, and the 2{,}704 of
% them outside the held-out validation split define the prototypes. The validation
% split is a protein-disjoint 10\% of the cached pool, 349 proteins and 2{,}467
% images, held out from every fit and never overlapping the test proteins.
%
%
%%%%%%%
During FACET module adaptation and semantic memory construct, we use a random subset of the
HPA training split, formed by adding whole proteins in a fixed random order until an image
budget is reached and keeping every image of each
selected protein. This yields 24{,}001 images from 3{,}491 of the 12{,}732 training proteins
($\approx$27\% of training images); the pretrained backbone, in contrast, was trained on the
full split. A protein-disjoint 10\% of this pool (349 proteins, 2{,}467 images) forms a
validation split that is excluded from every fit, is used only for hyperparameter tuning, and never overlaps the test proteins. Of the pool, 3{,}007 proteins
carry a UniProt localization annotation; the 2{,}704 of them outside the validation split
define the prototypes. 
Unannotated proteins are used only for the context map, the ridge and the
projections; prototype construction, prototype prior and posterior training,
adapter pretraining, transport training, and negative sampling all restrict to
annotated proteins.

%%%%%%%

For the HPA experiments, we evaluate four cellular-context regimes formed from
different combinations of the three available channels: nucleus, nucleus+ER,
nucleus+MT, and nucleus+ER+MT as done in prior work~\citep{zheng2025celldiff}.

\subsection{OpenCell benchmark}
\label{app:opencell}

The OpenCell benchmark is derived from endogenous-tagging microscopy
\citep{cho2022opencell}. {The released OpenCell training split contains 5,828 images from 1,211 proteins and the test split contains 473 stored
examples from 100 proteins}\footnote{\url{https://virtualcellmodels.cziscience.com/dataset/opencell-microscopy-images}}. Applying the same maximum tokenized sequence length of 2048 removes
19 examples, leaving 454 images from 97 proteins.

% \revnew{tell about val split a bit}

The OpenCell training split used for {OpenCell adaptation}
contains 5,659
images from 1,179 proteins after applying the same sequence-length filter. From these 10\% samples are held-out as validation set, containing 117 proteins and 544 corresponding images. 
OpenCell test proteins are also strictly disjoint from the HPA training split used to fit the semantic memory.
% and from the 3,491-protein pool used for FACET{} memory construction.

For OpenCell, we use the nucleus image as the cellular-context input. We
{evaluate HPA-module transfer of the HPA-trained \FACET{}
modules and further OpenCell adaptation of those modules on the OpenCell
training split.}

\subsection{Backbone-matched instantiation}
\label{app:backbone-matched}

\paragraph{Choice of backbone.} 
We instantiate \FACET{} on CELL-Diff because, to our knowledge,
it is the only publicly available sequence-conditioned latent diffusion model for this task with released
checkpoints for both HPA and OpenCell. Other recent methods, such as CELL-E2 \citep{khwaja2023celle2} and Vermeer
\citep{kambhampati2026vermeer} are autoregressive; PUPS \citep{zhang2025prediction} is a
deterministic single-cell inpainting model, for which distributional metrics and sampling-based
calibration are not defined; and ProtiCelli \citep{proticelli} conditions on protein identities
from a fixed vocabulary, rather than protein sequences, and therefore cannot be evaluated on proteins held out from training.

\paragraph{Instantiation.} For both HPA and OpenCell, the pretrained $\varepsilon$-predictor, VAE, and ESM-2 protein
language model \citep{lin2023esmfold} are taken from the corresponding dataset-specific
CELL-Diff checkpoint \citep{zheng2025celldiff}. The CELL-Diff baseline uses the same checkpoint
without modification. Sharing these components ensures that differences between CELL-Diff,
\FACET{}, and its internal variants reflect the conditioning design rather than differences in
backbone capacity or pretraining. This initialization is a choice of experimental design, not
part of \FACET{} itself, which applies to any pretrained variance-preserving
$\varepsilon$-predictor (Proposition~2).

All original backbone weights remain frozen. In variants with backbone LoRA, trainable
rank-16 matrices are inserted into 90 image-attention projections. The $\varepsilon$-adapter
and prototype prior $p_\eta$ are pretrained in Stage~I and then frozen, while the LoRA
parameters, hierarchical correction head, and continuous prior $p_\psi$ are optimized during
transport training (Appendix~\ref{app:training}). The posterior modules $q_\omega$ and $q_\phi$ are
used only during training and are not required for generation. In total, the deployed modules
add 9.83M trainable parameters (0.588\% of the 1.67B-parameter instantiation;
Table~\ref{tab:facet-parameters}).

\begin{table}[b]
\caption{Parameter counts for the trainable \FACET{} modules. Deployed modules are used during
generation; posterior modules are used only during training.}
\label{tab:facet-parameters}
\centering
\small
\setlength{\tabcolsep}{6pt}
\begin{tabular}{@{}lrc@{}}
\toprule
Module & Parameters & Deployed \\
\midrule
$\varepsilon$-adapter & 1,318,533 & \checkmark \\
Prototype prior $p_\eta$ & 923,145 & \checkmark \\
Hierarchical correction head & 4,918,340 & \checkmark \\
LoRA (rank 16) & 2,406,400 & \checkmark \\
Continuous prior $p_\psi$ & 262,785 & \checkmark \\
\midrule
\textbf{Total deployed} & \textbf{9,829,203} & -- \\
\midrule
Continuous posterior $q_\phi$ & 164,481 & -- \\
Prototype posterior $q_\omega$ & 3,020,297 & -- \\
\bottomrule
\end{tabular}
\end{table}

The LoRA rank, number of adapted projections, module depths, and optimization and sampling
constants specify the evaluated implementation; they are not claimed to be optimal choices.

\subsection{Hardware specifications}
\label{app:hardware}

All experiments were conducted on a compute node with eight CPU cores from an
AMD EPYC 9354 processor running at 3.25 GHz, 64 GB of system memory, and a
single NVIDIA L40S GPU with 48 GB of GDDR6 memory.

% ============================================================
\section{Metric Definitions and Evaluation Protocols}
\label{app:metrics}
% ============================================================

\subsection{{Image-quality metrics}}
\label{app:image-metrics}

We report mutual information (MI), multiscale structural similarity
(MS-SSIM), Haralick texture agreement, corpus-level peak signal-to-noise ratio
(PSNR), continuous intensity overlap
($\mathrm{IoU}_{\mathrm{cont}}$), and Fr\'echet inception distance (FID).
% \textcolor{orange}{
% Unless otherwise specified, MI, MS-SSIM, and Haralick
% agreement use the protein-fluorescence G channel of the stored RGB panel,
% normalized from uint8 to $[0,1]$, on the full $256\times256$ frame without
% masking or resizing. These scores are computed per target-generated pair,
% averaged within protein, and then macro-averaged across proteins.}
% \textcolor{orange}{TODO: Confirm and state explicitly whether PSNR and
% $\mathrm{IoU}_{\mathrm{cont}}$ use this same protein-fluorescence channel and
% preprocessing.}
% \textcolor{orange}{For the protein-macro image metrics, the HPA evaluation
% contains 696 images from 97 proteins and the OpenCell evaluation contains 454
% images from 97 proteins.}

\paragraph{Mutual information.}
MI measures statistical dependence between target and generated image
intensities. For target image intensities $X$ and generated image intensities
$\widehat{X}$, it is defined as
\begin{equation}
I(X;\widehat{X})
=
\sum_{x,\widehat{x}}
p(x,\widehat{x})
\log
\frac{p(x,\widehat{x})}{p(x)p(\widehat{x})}.
\label{eq:app-mi}
\end{equation}
% \textcolor{orange}{We digitize both images into 256 fixed intensity levels and compute raw mutual information in nats with \texttt{sklearn.metrics.mutual\_info\_score}, using the same estimator for every method without normalization or adjustment. The resulting pairwise scores are aggregated at the protein level before averaging across proteins.}

\paragraph{Multiscale SSIM.}
Following the multiscale SSIM formulation of~\cite{wang2003multiscale} and the fluorescence-image evaluation protocol of~\cite{proticelli}, we compute five-scale MS-SSIM on images normalized to \([0,1]\), using an \(11\times11\) uniform window and \(2\times2\) average pooling between scales. At scales 1-4, the SSIM map is spatially averaged before combining the scales. The same implementation is used for all methods before protein-level macro aggregation.
% \textcolor{orange}{
% MS-SSIM evaluates structural similarity across multiple
% spatial scales. We use a five-scale implementation with an $11\times11$
% uniform window, $2\times2$ average pooling between scales,
% $\texttt{data\_range}=1$, and weights
% $[0.0448,0.2856,0.3001,0.2363,0.1333]$. At scales 1-4, the contrast-structure
% term is the mean of the full SSIM map. These settings follow the fluorescence
% image-generation evaluation configuration of \citep{proticelli}; the same
% implementation is used for all methods before protein-level macro aggregation.}

\paragraph{Haralick texture agreement.}
We quantify texture similarity using six Haralick descriptors (contrast,
dissimilarity, homogeneity, energy, correlation, and angular second moment) computed
from gray-level co-occurrence matrices at a unit pixel offset and four orientations,
following~\cite{haralick1973textural}. We average each descriptor
over orientations and define texture agreement as the mean of the corresponding
per-feature relative similarities between real and generated images~\citep{proticelli}. 
% The same preprocessing is used for all methods, followed by protein-level macro aggregation.

% \paragraph{Haralick texture agreement.}
% \textcolor{orange}{Haralick agreement compares six texture properties extracted
% with 256 gray levels, distance 1, angles $0^\circ$, $45^\circ$, $90^\circ$,
% and $135^\circ$, using symmetric and normalized co-occurrence matrices. We
% average each property over angles and compute agreement as the mean relative
% similarity $1-|f_k-\widehat f_k|/\max(|f_k|,|\widehat f_k|,10^{-10})$. The same
% preprocessing is applied to every method, and scores are aggregated at the
% protein level.}

\paragraph{Peak signal-to-noise ratio.}
For corpus-level mean squared error $\mathrm{MSE}_{\mathrm{corpus}}$ and
maximum image intensity $L$, PSNR is
\begin{equation}
\mathrm{PSNR}
=
10\log_{10}
\left(
\frac{L^2}{\mathrm{MSE}_{\mathrm{corpus}}}
\right).
\label{eq:app-psnr}
\end{equation}
{All normalized image intensities use $L=1$.}

\paragraph{Continuous intensity overlap.}
For nonnegative target and generated fluorescence-intensity maps
$Y$ and $\widehat{Y}$, we compute the threshold-free weighted Jaccard
similarity, also known as the generalized Jaccard coefficient or Ruzicka
similarity:
\begin{equation}
\mathrm{IoU}_{\mathrm{cont}}(Y,\widehat{Y})
=
\frac{\sum_u \min\{Y_u,\widehat{Y}_u\}}
     {\sum_u \max\{Y_u,\widehat{Y}_u\}},
\end{equation}
where $u$ indexes spatial locations. We compute this score for each
target-generated pair, average scores within each protein, and then
macro-average across proteins.

\paragraph{Fr\'echet Inception Distance.}
We measure distributional similarity using the Fréchet Inception Distance
(FID)~\citep{heusel2017gans}. Each evaluation panel is represented as a
three-channel image whose red channel is zero, green channel is the
protein-fluorescence image, and blue channel is the corresponding nucleus
image~\citep{zheng2025celldiff,khwaja2023celle2}. We extract 2048-dimensional pool3 features from Inception-v3 and
compute FID 
% at the dataset level 
using the same panel construction and
feature-extraction pipeline for every method. Let
$(\mu_r,\Sigma_r)$ and $(\mu_g,\Sigma_g)$ denote the empirical means and
covariances of the real and generated feature embeddings. Then
\begin{equation}
\mathrm{FID}
=
\|\mu_r-\mu_g\|_2^2+
\operatorname{Tr}\!\left(
\Sigma_r+\Sigma_g-2(\Sigma_r\Sigma_g)^{1/2}
\right).
\end{equation}
Lower values indicate closer agreement between the real and generated
image distributions.
% \paragraph{Fr\'echet inception distance.}
% FID compares target and generated image distributions in a fixed feature
% space. \textcolor{orange}{For each stored RGB panel, the red channel is zero,
% the green channel is the protein-fluorescence image, and the blue channel is
% the real nucleus image. We use the 2048-dimensional final-pool feature block
% of the TF-ported InceptionV3 network from \texttt{pytorch-fid} 0.3.0, including
% its bilinear resize from $256\times256$ to $299\times299$ and normalization
% from $[0,1]$ to $[-1,1]$.} Let $(\mu_r,\Sigma_r)$ and $(\mu_g,\Sigma_g)$ denote the empirical means
% and covariances of target and generated feature embeddings. We compute
% \begin{equation}
% \mathrm{FID}
% =
% \|\mu_r-\mu_g\|_2^2
% +
% \operatorname{Tr}
% \left(
% \Sigma_r+\Sigma_g
% -
% 2(\Sigma_r\Sigma_g)^{1/2}
% \right).
% \label{eq:app-fid}
% \end{equation}
% FID is computed at the dataset level. Lower values indicate closer agreement
% between the target and generated image distributions.

\paragraph{Network function evaluations.}
NFE denotes the number of learned-model evaluations required during sampling and measures sampling cost. The primary FACET operating point uses 25 NFE, while the ablations consider operating points down to 10 NFE. The backbone-matched baseline uses 100 NFE.

\subsection{Protein-association evaluation}
\label{app:association-metric}

For the protein-association analysis, we follow the protocol introduced by~\cite{proticelli}. We use STRING
\citep{szklarczyk2023string} solely to obtain external evaluation labels.
For each protein pair, we compute within-cell Pearson similarity after
removing the per-cell mean image across proteins. This removes similarity
caused by shared cellular morphology.

{For each system and STRING threshold, we compare
STRING-associated pairs with localization-matched non-associated pairs when
both proteins share a primary compartment; otherwise, we use all
non-associated pairs as the unmatched null.} We report
\begin{equation}
\Delta
=
\mathbb{E}
\left[
r(a,b)\mid\mathrm{associated}
\right]
-
\mathbb{E}
\left[
r(a,b)\mid\mathrm{matched\ null}
\right],
\label{eq:app-association-delta}
\end{equation}
where $r(a,b)$ is the within-cell Pearson similarity. Positive values indicate
stronger association-consistent spatial patterns.

Statistical significance tests compare associated pairs with their corresponding null
pairs and are assessed separately within
each system using a two-sided Mann-Whitney U test \citep{mann1947test}.
% These tests compare associated pairs with their corresponding null pairs; they
% do not test the difference between FACET and CELL-Diff.

\subsection{Predictive calibration metrics}
\label{app:calibration-metrics}

We evaluate the expected calibration error (ECE),
maximum calibration error (MCE), and the Brier score~\citep{guo2017calibration,glenn1950verification}.
For each input condition, stochastic generation produces an empirical
foreground probability $p_u$ at spatial location $u$. Let
$\mathcal{B}_1,\ldots,\mathcal{B}_M$ denote fixed probability bins. Expected
calibration error (ECE) is
\begin{equation}
\mathrm{ECE}
=
\sum_{m=1}^{M}
\frac{|\mathcal{B}_m|}{N}
\left|
\operatorname{acc}(\mathcal{B}_m)
-
\operatorname{conf}(\mathcal{B}_m)
\right|,
\label{eq:app-ece}
\end{equation}
where $\operatorname{acc}(\mathcal{B}_m)$ is the empirical foreground
frequency and $\operatorname{conf}(\mathcal{B}_m)$ is the mean predicted
probability in bin $m$.

Maximum calibration error (MCE) is
\begin{equation}
\mathrm{MCE}
=
\max_m
\left|
\operatorname{acc}(\mathcal{B}_m)
-
\operatorname{conf}(\mathcal{B}_m)
\right|.
\label{eq:app-mce}
\end{equation}

The Brier score is
\begin{equation}
\mathrm{Brier}
=
\frac{1}{N}
\sum_{u=1}^{N}
(p_u-y_u)^2,
\label{eq:app-brier}
\end{equation}
where $y_u$ is the binary foreground label. Lower ECE and
MCE indicate smaller calibration gaps, whereas lower Brier score indicates
better overall probabilistic forecast quality.
% ; the Brier score is not a pure calibration error.

% ============================================================
\section{Statistical and Sampling Protocol}
\label{app:statistical-protocol}
% ============================================================

All methods are evaluated on identical held-out examples, protein sequences,
cellular-context inputs, and target images for each comparison. Unless
otherwise specified, the reported image metrics use the same preprocessing,
feature extraction, aggregation rules, and checkpoint across methods.

\subsection{Predictive calibration protocol}
\label{app:calibration-protocol}

The calibration experiment uses a strictly held-out fitting partition rather
than test-set fitting or protein-wise cross-validation. We generate a second
predictive grid on a held-out training-validation set containing 96 images from
81 proteins. This partition shares zero proteins, image identifiers, or
storage keys with the 696-image, 97-protein HPA test benchmark.

Both partitions use the identical checkpoint, sampler, 25-NFE operating point,
stochasticity scale, adapter and hierarchical-head settings, and forecast
reduction procedure.

For each \FACET{} image, four residual-latent draws are crossed with four
stochastic-differential-equation noise seeds, producing 16 predictive samples.
The backbone-matched reference uses the same total predictive-sample
budget. {Before test scoring, we determine the foreground
threshold from the 349 held-out validation proteins as}
\begin{equation}
{
t=\operatorname{median}_{p\in\mathcal V}
\left[
\frac{1}{|\mathcal I_p|}
\sum_{i\in\mathcal I_p}
Q_{0.99}\!\left(\{v_{ij}:j\in B_i\}\right)
\right],
}
\label{eq:pred_cal_threshold}
\end{equation}
{where $\mathcal V$ is the set of held-out validation
proteins, $\mathcal I_p$ is the set of validation images for protein $p$,
$v_{ij}$ is the intensity at pixel $j$ in image $i$, and $B_i$ is the valid
image region excluding the dilated cell mask. This gives
$t=6.111$ on the stored $0$--$255$ uint8 scale
($t/255\approx0.02397$). The threshold is then held fixed for test
evaluation. 
For HPA calibration, each sample is binarized using this threshold,
and the empirical foreground probability is defined from the 16 binarized
samples:}
\[
{
p_u=\frac{1}{16}\sum_{s=1}^{16}\mathbf{1}[x_s(u)>t].
}
\]
{Here $x_s(u)$ is the intensity at location $u$ in sample
$s$. Calibration uses 15 equal-width bins on $[0,1]$, with the final bin
including the upper endpoint; the Brier score is computed directly from the
pixel-level probabilities and labels~\citep{guo2017calibration}.}

{For post-hoc calibration, we follow the temperature-scaling
procedure introduced by \cite{guo2017calibration}.} 
Note that the threshold in Eq.~\ref{eq:pred_cal_threshold} is computed over all 349 held-out validation proteins, whereas the temperature is fit on the 96-image, 81-protein subset of those proteins for which the 16-sample predictive grid was generated.
Because raw sample proportions can
be exactly $0$ or $1$, we first use Jeffreys smoothing~\citep{jeffreys1946invariant,brown2001interval}
$p_a=(k+\alpha)/(16+2\alpha)$ with $k$ the number of foreground samples and
$\alpha=0.5$, transform to logit space, and apply
$p'=\sigma(\operatorname{logit}(p_a)/T)$, with probabilities clipped to
$[10^{-12},1-10^{-12}]$ when evaluating the negative log-likelihood. One scalar
temperature is fit for each model by validation negative log-likelihood and
then applied unchanged to that model's test forecasts. No calibration parameter
is estimated from test pixels or test proteins; Table~\ref{tab:res-calibration}
reports the post-hoc calibrated values.

\subsection{Sampling efficiency}
\label{app:sampling-efficiency}

The HPA operating point uses 75\% fewer network function evaluations than the
backbone-matched diffusion reference \citep{zheng2025celldiff}. On
matched hardware (Appendix~\ref{app:hardware}), the measured generation loop for the 696-image HPA
evaluation set takes approximately 15 minutes for \FACET{} and 78 minutes for
the reference, corresponding to an approximately 80.8\% reduction in
wall-clock time.

Because context loading and other system-level operations may introduce small
run-to-run variations, NFE is used as the primary compute measure. The
wall-clock measurements are reported as a complementary hardware-specific
observation.

% ============================================================
\section{Additional Analyses}
\label{app:additional-analyses}
% ============================================================

\subsection{Conditioning sensitivity}
\label{app:conditioning-sensitivity}

We estimate the local sensitivity of the learned correction to $z_S$ and
$z_C$ using Hutchinson trace estimators~\citep{hutchinson1989stochastic}
of
$\operatorname{tr}(J_{z_a}J_{z_a}^{\mathsf T})=\|J_{z_a}\|_F^2$,
where $J_{z_a}=\partial\Delta v/\partial z_a$ is the Jacobian of the learned
velocity correction with respect to $z_a$, for $a\in\{S,C\}$. We evaluate
20 images from 18 proteins and five timesteps.

In the implemented latent parameterization, the median raw ratio
\[
\frac{\|J_{z_S}\|_F}{\|J_{z_C}\|_F}
\]
is 28.0, with 10th and 90th percentiles of 13.0 and 48.3, respectively.
{Because this ratio uses unnormalized Jacobian norms and
$z_S$ and $z_C$ have different latent scales, we treat it as a descriptive
diagnostic rather than a scale-independent comparison. To account for the
observed scale of each latent, we additionally report the normalized
sensitivities}
\[
{
\rho_{z_a}
=
\frac{\|J_{z_a}\|_F\|z_a\|_2}{\|\Delta v\|_F},
\qquad a\in\{S,C\},
}
\]
{where $\Delta v$ is the learned velocity correction. The
median normalized sensitivities are 3.645 for $z_S$ and 0.606 for $z_C$,
with 10th and 90th percentiles of $[1.236,9.638]$ and $[0.431,1.157]$,
respectively. Thus, under the observed latent scales, the median normalized
sensitivity to $z_S$ is approximately sixfold that to $z_C$.}

This diagnostic measures local operating sensitivity of the learned correction.
It does not establish a causal benefit of the FiLM ordering because no
reversed-order model was trained. Instead, it provides supporting evidence that
the protein and cellular-context pathways have distinct and measurable effects
within the implemented hierarchical conditioning mechanism.

\subsection{Sampling and schedule representability}
\label{app:flow}

For the reported uniform Euler-Maruyama implementation, integration uses the
fixed range $\tau\in[0.05,1]$. The lower cutoff is an implementation choice
associated with the inherited finite native schedule and numerical endpoint
handling; it is not a universal \FACET{} constant.

With $n$ evaluations on the uniform grid $\tau_j=\tau_{\min}+j\Delta\tau$, $j=0,\dots,n-1$, the
final (lowest-noise) evaluation occurs at
\begin{equation}
\tau_{\mathrm{emit}}=\tau_{\max}-\frac{\tau_{\max}-\tau_{\min}}{n}=1-\frac{0.95}{n}.
\end{equation}
Each native noise level $k$ corresponds to the unique coordinate $\tau_k$ satisfying
$a_{\tau_k}=\sqrt{\bar\alpha_k}$, since $a_\tau$ is strictly increasing. The least-noisy level of the pretrained predictor's 200-level cosine training schedule
($\bar\alpha\approx0.99975$) maps to $\tau\approx0.98588$. We call an operating point supported if every evaluation lies within the
range spanned by the native schedule, so that nearest-index lookup never falls back to the
boundary level. Requiring $\tau_{\mathrm{emit}}\le0.98588$ gives $n\le67.3$, i.e.,
$\mathrm{NFE}\le67$: at NFE 67, $\tau_{\mathrm{emit}}=0.985821$, whereas at NFE 68,
$\tau_{\mathrm{emit}}=0.986029$ falls outside this range. The lower cutoff $\tau_{\min}=0.05$ lies
well within the native range.

% With $n$ evaluations, the first emitted high-noise state is
% \begin{equation}
% \tau_{\mathrm{emit}}
% =
% 1-\frac{0.95}{n}.
% \label{eq:app-tau-emit}
% \end{equation}
% The largest native schedule value is 0.98588. Requiring
% $\tau_{\mathrm{emit}}\leq 0.98588$ gives
% \[
% n\leq67.3.
% \]
% Therefore, under this specific schedule mapping, the supported integer
% operating range satisfies
% \[
% \mathrm{NFE}\leq67.
% \]
% At NFE 67, $\tau_{\mathrm{emit}}=0.985821$, whereas at NFE 68,
% $\tau_{\mathrm{emit}}=0.986029$, which lies beyond the supported range.

% This bound is specific to the reported discretization and inherited native
% schedule. It is not a universal upper bound for flow matching or stochastic
% transport.
% {The reported 25-NFE operating point lies within this
% implementation-specific supported range.}

% ============================================================
\section{Additional Qualitative Results}
\label{app:additional-qualitative}
% ============================================================

The following figures complement the qualitative comparison presented in the
main paper. They use the same visualization conventions and method names as
the main results. 
% Per-example $\mathrm{IoU}_{\mathrm{cont}}$ values are omitted because the aggregate quantitative results already report this metric and are more appropriate for dataset-level comparison.

\subsection{{Cellular-context ablation}}
\label{app:qual-context-ablation}

\begin{figure}[t]
\centering
\IfFileExists{figures/fig_hpa_context_ablation_band.pdf}{%
    \includegraphics[width=\columnwidth]{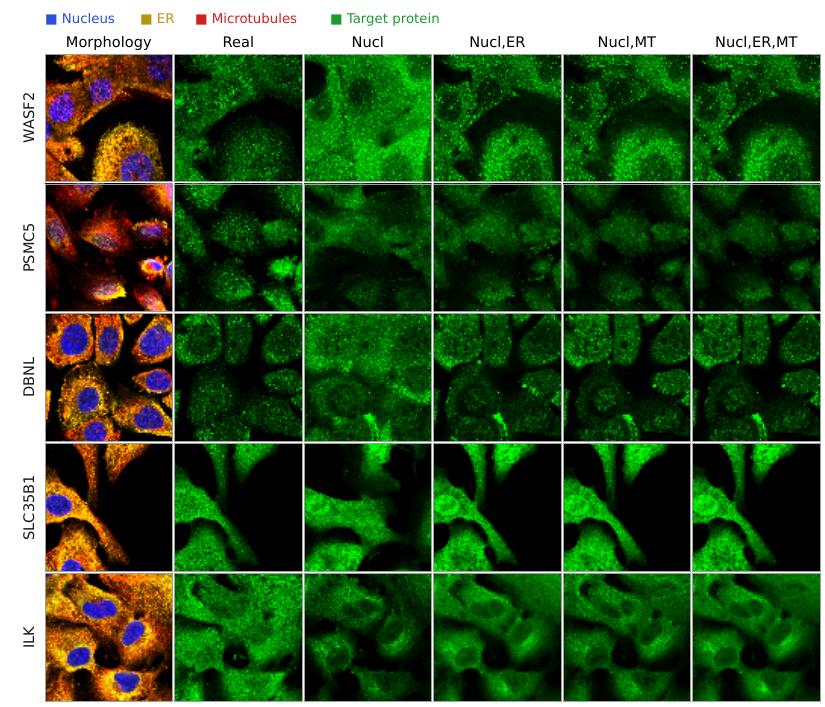}%
}{%
    \fbox{\parbox[c][0.24\textheight][c]{0.94\columnwidth}{%
    \centering
    \textcolor{orange}{[TODO: insert final cellular-context ablation figure]}\\
    Cellular-context ablation
    }}%
}
\caption{{Cellular-context ablation. The protein sequence is
held fixed while the cellular-context input is removed or modified. The
examples illustrate context-conditioned changes in the generated spatial
phenotype; they are qualitative evidence rather than a causal ablation beyond
the quantitative comparison reported in the main text.}}
\label{fig:app-context-ablation}
\end{figure}

{Figure~\ref{fig:app-context-ablation} visualizes the
cellular-context ablation. The protein sequence is held fixed while the
combination of cellular-context channels is altered.}

\subsection{Same-protein generation across different contexts}
\label{app:qual-hpa-same-protein}

\begin{figure}[t]
\centering
\IfFileExists{figures/fig_ctx_fixed_protein_hpa.pdf}{%
    \includegraphics[width=.48\columnwidth]{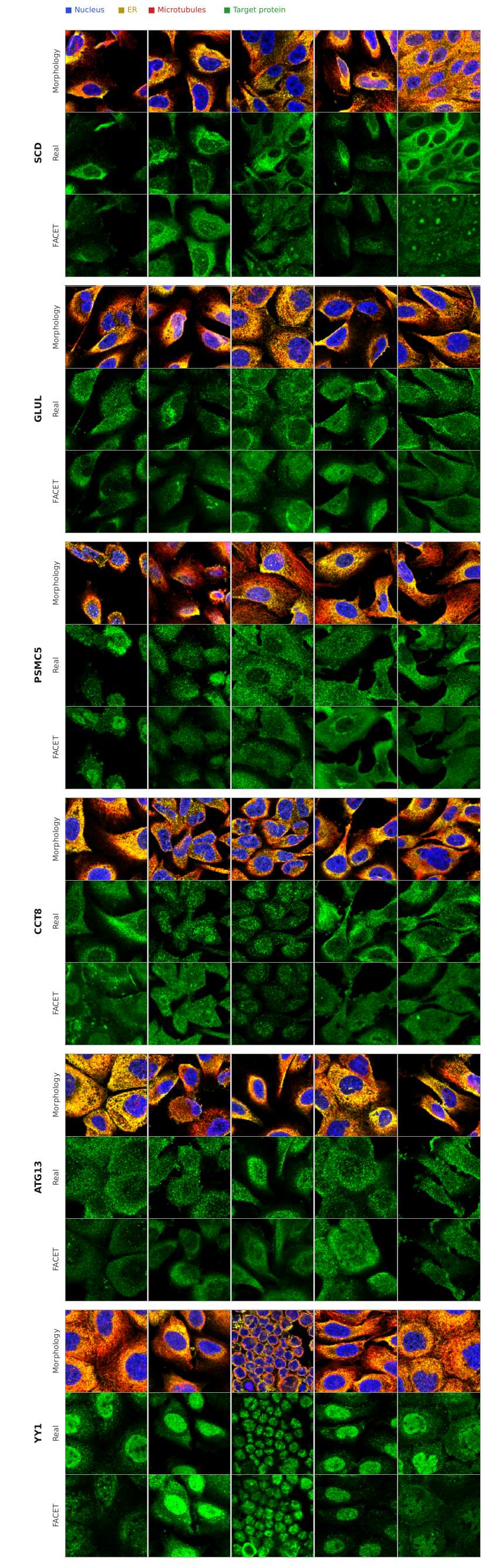}%
}{%
    \fbox{\parbox[c][0.24\textheight][c]{0.94\columnwidth}{%
    \centering
    \textcolor{orange}{[TODO: insert final HPA same-protein figure]}\\
    HPA same-protein, different-context examples
    }}%
}
\caption{{Same-protein generation across HPA cellular
contexts. The sequence input is fixed across contexts, allowing
context-conditioned changes in the generated spatial phenotype to be
visualized.}}
\label{fig:app-hpa-same-protein}
\end{figure}

Figure~\ref{fig:app-hpa-same-protein} provides additional examples in
which the same protein sequence is evaluated under distinct cellular contexts.

\subsection{Additional protein and context combinations on HPA}
\label{app:qual-hpa-grid}

\begin{figure*}[t]
\centering
\IfFileExists{figures/fig_ctx_fixed_context.pdf}{%
    \includegraphics[width=\textwidth]{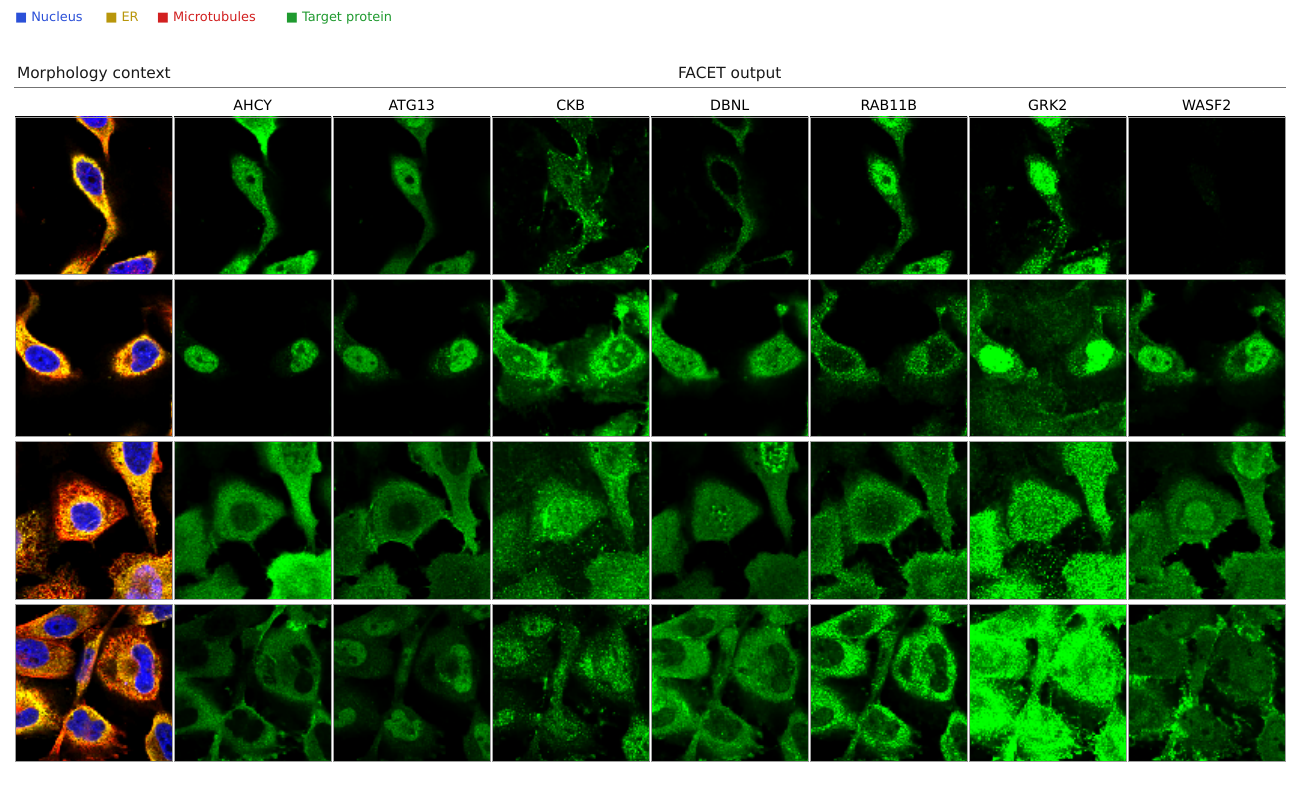}%
}{%
    \fbox{\parbox[c][0.24\textheight][c]{0.94\textwidth}{%
    \centering
    \textcolor{orange}{[TODO: insert final HPA protein/context grid]}\\
    HPA protein/context generation grid
    }}%
}
\caption{{Additional HPA generations across multiple proteins
and cellular contexts. The examples provide broader coverage of the
conditional-generation setting beyond the main-text qualitative comparison.}}
\label{fig:app-hpa-grid}
\end{figure*}

Figure~\ref{fig:app-hpa-grid} shows a broader grid of different proteins and
cellular contexts on HPA. This figure provides broader qualitative coverage
beyond the representative examples shown in the main paper.

\subsection{Additional qualitative comparison on OpenCell}
\label{app:qual-opencell-comparison}

\begin{figure*}[t]
\centering
\IfFileExists{figures/facet_qualitative_opencell6.pdf}{%
    \includegraphics[width=\textwidth]{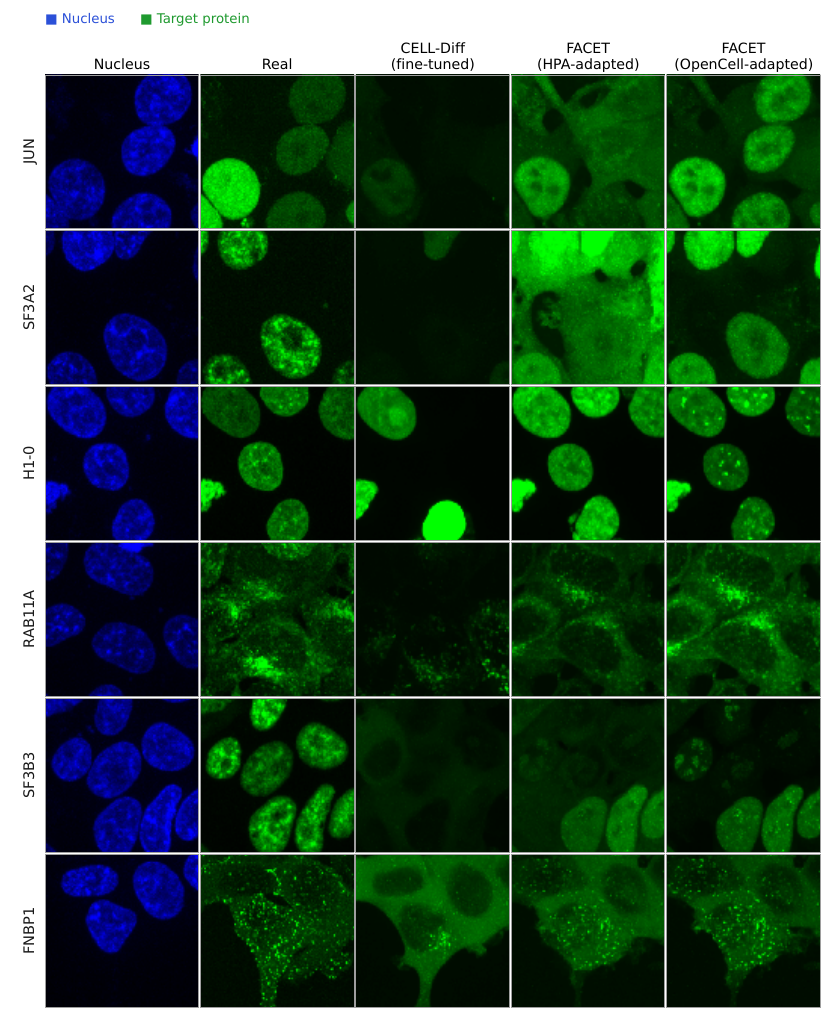}%
}{%
    \fbox{\parbox[c][0.24\textheight][c]{0.94\textwidth}{%
    \centering
    \textcolor{orange}{[TODO: insert final OpenCell qualitative comparison]}\\
    Additional OpenCell qualitative comparison
    }}%
}
\caption{Additional qualitative results on OpenCell. The
figure uses the same method names and visualization conventions as the main
OpenCell results and complements the quantitative comparison. Here ``CELL-Diff (fine-tuned)'' represents the CELL-Diff checkpoint fine-tuned on OpenCell data, the same one reported in the main text and the source~\citep{zheng2025celldiff}.}
\label{fig:app-opencell-comparison}
\end{figure*}

Figure~\ref{fig:app-opencell-comparison} presents additional qualitative
comparisons on OpenCell. 
% The columns should use the same terminology as the main quantitative and qualitative results:
% \[
% \text{Context}
% \;|\;
% \text{Target}
% \;|\;
% \text{CELL-Diff}
% \;|\;
% \text{\FACET{} (HPA-adapted)}
% \;|\;
% \text{\FACET{} (OpenCell-adapted)}.
% \]

\subsection{Additional qualitative comparison on HPA}
\label{app:qual-hpa-comparison}

\begin{figure*}[t]
\centering
\IfFileExists{figures/facet_qualitative_appendix4.pdf}{%
    \includegraphics[width=\textwidth]{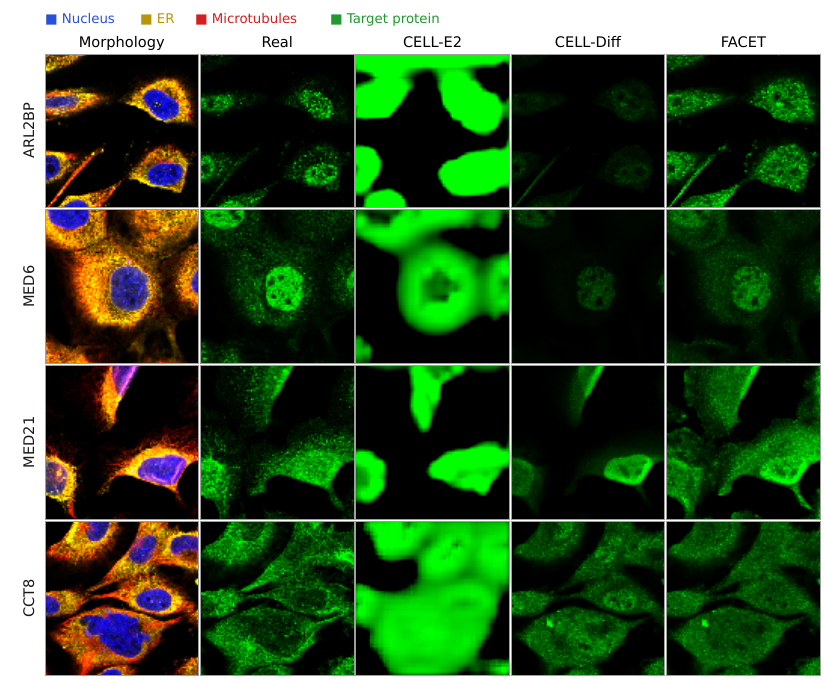}%
}{%
    \fbox{\parbox[c][0.24\textheight][c]{0.94\textwidth}{%
    \centering
    \textcolor{orange}{[TODO: insert final HPA qualitative comparison]}\\
    Additional HPA qualitative comparison
    }}%
}
\caption{{Additional qualitative results on HPA. CELL-E2
uses nucleus-only context, whereas CELL-Diff and FACET use the full
three-channel context. These examples extend the main-text qualitative
comparison using the same evaluation setting, method names, and visualization
conventions.}}
\label{fig:app-hpa-comparison}
\end{figure*}

Figure~\ref{fig:app-hpa-comparison} provides additional HPA qualitative
comparisons beyond the main-text qualitative figure. It uses the same method
ordering and visualization conventions as the main comparison.

\section{Disclosure of Generative AI Use}
Large Language Models (LLMs) were \textbf{used only to polish the text}; they \textbf{were not involved} in retrieval, discovery, or research ideation.

\end{document}